%% file: main.tex
\documentclass{article}

\PassOptionsToPackage{numbers,sort&compress}{natbib}

\usepackage[main,final]{neurips_2026}

\usepackage[utf8]{inputenc} %
\usepackage[T1]{fontenc}    %
\usepackage{hyperref}       %
\usepackage{url}            %
\usepackage{booktabs}       %
\usepackage{amsfonts}       %
\usepackage{amsmath}        %
\usepackage{amssymb}        %
\usepackage{nicefrac}       %
\usepackage{microtype}      %
\usepackage{xcolor}         %
\usepackage{enumitem}       %
\usepackage{algorithm}      %
\usepackage{algorithmic}    %
\usepackage{bbm}            %
\usepackage{makecell}       %
\usepackage{graphicx}       %
\usepackage[caption=false]{subfig} %
\usepackage[nohints]{minitoc} %
\usepackage{multicol}
\usepackage{multirow}

\usepackage{pifont}\usepackage{fontawesome5}
\newcommand{\xmark}{\ding{55}}
\newcommand{\cmark}{\ding{51}}

\usepackage{xspace}

\newcommand{\longcounsel}{\textsc{CounselArc}\xspace}
\newcommand{\daic}{\textsc{Daic-Woz}\xspace}
\newcommand{\emotrackCodeURL}{https://github.com/Xtra-Computing/EmoTrack}

\title{EmoTrack: Clinical-Semantic Modeling for Text-Based Depression Severity Estimation}

 \author{%
  Zhaomin Wu,\;Jiayi Li,\;Bingsheng He \\
  Department of Computer Science\\
  National University of Singapore\\
  \texttt{zhaomin@nus.edu.sg, li.jiayi@u.nus.edu, dcsheb@nus.edu.sg}
}

\begin{document}

\doparttoc
\faketableofcontents

\maketitle

\begin{abstract}
\input{section/abstract.tex}

\end{abstract}\begingroup\renewcommand{\thefootnote}{}\footnotetext[0]{\faGithub\ Code: \href{\emotrackCodeURL}{github.com/Xtra-Computing/EmoTrack}\quad \faDatabase\ Dataset: \href{https://huggingface.co/datasets/hiddensev/LongCounsel-8}{huggingface.co/datasets/hiddensev/LongCounsel-8}}\endgroup

\input{section/introduction.tex}
\input{section/related_work.tex}
\input{section/problem-statement.tex}
\input{section/approach.tex}

\input{section/phq_dataset.tex}
\input{section/dataset_validation.tex}
\input{section/experiment.tex}

\input{section/conclusion}

\begin{ack}
\label{sec:ack}
We thank Adrian Toh, Clinical Psychologist, for his valuable clinical insights and expert guidance in clinical psychology, particularly in informing the design and development of the counselling dataset and depression severity estimation pipeline.

This research is supported by the National Research Foundation, Singapore and Infocomm Media Development Authority under its Trust Tech Funding Initiative. Any opinions, findings and conclusions or recommendations expressed in this material are those of the authors and do not reflect the views of National Research Foundation, Singapore and Infocomm Media Development Authority. This research is also in part supported by Zhaomin Wu's National Research Foundation (NRF) Postdoctoral Award.
\end{ack}

{
\small
\bibliographystyle{plainnat}
\bibliography{references}
}

\clearpage
\addcontentsline{toc}{section}{Appendix}
\noindent\begin{minipage}{\linewidth}
\hypersetup{hidelinks}
\part{Appendix}
\parttoc
\end{minipage}\par

\appendix

\input{section/add_related_work.tex}

\input{section/add_algorithm}

\input{section/add_phq_dataset}

\input{section/add_experiment_setting.tex}

\input{section/add_experiment.tex}

\input{section/limitation}
\input{section/license}

\clearpage
\input{checklist.tex}

\end{document}

%% file: section/abstract.tex
Text-based counseling is widely adopted across online mental health platforms, where automatically estimating depressive symptom severity from transcripts is vital for client safety and clinical monitoring. We therefore study prediction of the standard eight-item Patient Health Questionnaire (PHQ-8) total score directly from counseling sessions. Existing clinical-based methods obtain structured assessments through large language model inference but summarize away individual client turns. Training-based methods preserve turn semantics but must learn clinical structure from limited labels. Integrating these representations requires preserving clinical-feature identity and selecting evidence relevant to each symptom. We propose \textbf{EmoTrack}, which jointly encodes clinical-feature tokens and frozen client-turn embeddings, then uses PHQ-8 item queries to estimate total severity. Optional gated memory incorporates evidence from the preceding session. To support controlled evaluation in the multi-session setting, we introduce \longcounsel, a synthetic counseling dataset with session-level PHQ-8 supervision for predicting current-session severity with optional prior-session context. Across real \daic interviews, transfer to \textsc{E-DAIC}, and three longitudinal datasets, EmoTrack achieves the lowest normalized aggregate error. It reduces \daic mean absolute error by $12.5\%$ and aggregate error by approximately $8.6\%$, each relative to the strongest baseline for that measure.

%% file: section/introduction.tex
\section{Introduction}\label{sec:intro}

Text-based counseling has become increasingly prevalent as the primary mode of care across online mental health platforms~\citep{Liu2023ChatCounselorAL,qiu2025psydial,Xu2025MentalChat16KAB}. On these platforms, automatically estimating depressive symptom severity directly from session transcripts is critical for ensuring client safety, enabling timely risk detection and escalation. To ground this task in established clinical practice, we study the prediction of the widely adopted eight-item Patient Health Questionnaire (PHQ-8; scored $0$--$24$)~\citep{kroenke2009phq8} from counseling transcripts. Furthermore, because online counseling often spans multiple encounters, earlier sessions can provide valuable longitudinal context for assessing a client's current state.

Existing methods obtain clinical information through two complementary routes. \emph{Clinical-based methods} rely mainly on LLM inference to complete questionnaires or score clinician-style features, which are then mapped to severity estimates~\citep{rosenman2024llm,lee2026interpretable}. These assessments supply explicit clinical structure but summarize the session without retaining individual client turns as separate predictor inputs. \emph{Training-based methods} train predictors directly on transcripts or their semantic embeddings~\citep{lau2023automatic,milintsevich2023towards,ravenda2025transforming}. They preserve local conversational detail but must learn its relationship to symptom severity from scarce labeled transcripts. Combining prompted clinical features with turn-level semantics can bring clinical structure into transcript modeling under limited supervision.

This combination presents two challenges for single-session estimation and a further challenge for longitudinal use. \textbf{(i) Clinical scores and dialogue turns have mismatched structures.} Assessments form a small set of named scalars, whereas client turns form a variable-length sequence of text embeddings. Fusing them must retain what each score represents while connecting it to conversational evidence. \textbf{(ii) Different PHQ-8 items require different evidence.} A turn about insomnia may inform the sleep item, whereas feeling worthless may inform low self-worth. A pooled session vector forces all item predictors to rely on the same compressed evidence. \textbf{(iii) History may be missing or outdated.} Prior disclosures can clarify indirect statements but may mislead prediction after symptoms change. First or isolated sessions have no history.

To address these challenges, we build on the key idea of representing clinical assessments and client turns as distinct, retrievable evidence for each PHQ-8 item, and propose \textbf{EmoTrack} with three corresponding components. (i) To align these mismatched structures, we convert both inputs into a shared token space: clinical-feature tokens combine a score projection with feature- and group-identity embeddings, while client-turn tokens project frozen semantic embeddings for joint Transformer encoding. (ii) Instead of pooling a single session vector, we assign one learned query per PHQ-8 item to retrieve symptom-relevant evidence, whose resulting item predictions sum to the total score. (iii) To handle missing or shifting history, an optional memory module compresses the preceding session via symptom-specific retrieval and regulates its contribution with learned gates, leaving the current-session architecture intact when history is absent.

Beyond modeling, evaluating longitudinal symptom tracking faces a severe data bottleneck: it requires multi-session counseling dialogues paired with session-level symptom targets. However, real clinical interviews such as \daic only provide single-session PHQ-8 supervision~\citep{Gratch2014TheDA}, while existing multi-session counseling corpora focus on dialogue quality or continuity without aligned clinical labels~\citep{qiu2025psydial,pan2026psycheval}. To bridge this evaluation gap, we introduce \longcounsel, a synthetic longitudinal benchmark that grounds five-session counseling dialogues in client profiles, empirical severity trajectories, symptom compositions, and private behavioral cues, thoroughly audited for both label fidelity and dialogue realism.

Our contributions are threefold. (1) We introduce EmoTrack for joint clinical--semantic PHQ-8 estimation, with item-specific queries and optional previous-session memory. (2) We introduce \longcounsel, a controlled synthetic multi-session counseling dataset with prescribed symptom trajectories and session-level PHQ-8 supervision, supporting current-session severity prediction under partial symptom disclosure and cross-session continuity. (3) We evaluate on real \daic interviews, frozen transfer to \textsc{E-DAIC}, and three longitudinal datasets, with controlled representation, readout, objective, and memory ablations. EmoTrack reduces \daic MAE by $12.5\%$ relative to AIDA and achieves the lowest normalized aggregate error across the five datasets.

%% file: section/related_work.tex
\section{Related work}
\label{sec:related}

We organize transcript-based depression severity prediction by how clinical evidence enters the model. Clinical-based methods train downstream regressors on prompted assessments, while training-based methods learn from transcripts or turn embeddings.

\input{table/depression_track_studies_short}

\paragraph{Clinical-based methods.}
Prompted LLMs convert transcripts into structured questionnaire responses or clinician-style assessments. LMIQ completes questionnaires and supplies the coded responses to a random-forest regressor~\citep{rosenman2024llm}. AIDA scores 23 clinical, linguistic, and cognitive features and maps them to PHQ-8 severity with regression~\citep{lee2026interpretable}. These methods provide explicit clinical structure, but their session-level feature vectors do not retain individual client turns as separate predictor inputs. EmoTrack jointly represents prompted scores and client turns, enabling each PHQ-8 item query to access both sources of evidence.

\paragraph{Training-based methods.}
Supervised predictors learn from transcripts or turn embeddings under limited labeled data. Lau et al.~\citep{lau2023automatic} use parameter-efficient text-encoder tuning. Milintsevich et al.~\citep{milintsevich2023towards} predict PHQ-8 items hierarchically and sum their scores. PsychiatricPrism~\citep{agarwal2024analyzing} uses a transformer-based symptom predictor and studies psychiatrist annotations. EnsemBERT~\citep{ravenda2025transforming} predicts BDI-II items from social-media text and is adapted here to PHQ-8. These methods retain local text evidence, while EmoTrack additionally supplies prompted clinical-feature tokens and optional previous-session memory (Table~\ref{tab:lit-emotion-tracking-short}). Because interviewer prompts can provide shortcuts~\citep{burdisso2024daic}, we focus on client turns and test speaker-view alternatives.

\paragraph{Counseling data and longitudinal evaluation.}
\daic and \textsc{E-DAIC} provide real single-interview evaluation~\citep{Gratch2014TheDA,ringeval2019avec}. Counseling resources such as PsyDial, CACTUS, and PsychEval broaden dialogue coverage~\citep{qiu2025psydial,lee2024cactus,pan2026psycheval}, but do not provide the controlled five-session PHQ-8 trajectories needed here. We therefore introduce \longcounsel, pairing synthetic counseling trajectories with per-session symptom states and simulated self-reports. Section~\ref{sec:phq-dataset} describes its construction, and Appendix~\ref{sec:add_related} compares the wider method and dataset landscape. This resource complements real single-session evaluation with controlled longitudinal tests.

%% file: table/depression_track_studies_short.tex
\begin{table}[ht!]
  \centering
  \small
  \caption{Methods for text-based depression severity prediction. Clinical knowledge denotes prompted clinical features, and training denotes learning from transcripts or turn embeddings.}
  \label{tab:lit-emotion-tracking-short}
  \setlength{\tabcolsep}{3.5pt}
  \begin{tabular}{lcccc}
    \toprule
    \textbf{Study} & \textbf{Clinical knowledge} & \textbf{Training} & \textbf{Longitudinal} & \textbf{Target scale} \\
    \midrule
    Lau et al.~\cite{lau2023automatic} & \xmark & \cmark & \xmark & PHQ-8 \\
    Milintsevich et al.~\cite{milintsevich2023towards} & \xmark & \cmark & \xmark & PHQ-8 \\
    PsychiatricPrism~\cite{agarwal2024analyzing} & \xmark & \cmark & \xmark & PHQ-8 \\
    EnsemBERT~\cite{ravenda2025transforming} & \xmark & \cmark & \xmark & BDI-II \\
    LMIQ~\cite{rosenman2024llm} & \cmark & \xmark & \xmark & PHQ-8 \\
    AIDA~\cite{lee2026interpretable} & \cmark & \xmark & \xmark & PHQ-8 \\
    \midrule
    \textbf{EmoTrack (ours)} & \cmark & \cmark & \cmark & PHQ-8 \\
    \bottomrule
  \end{tabular}
\end{table}

%% file: section/problem-statement.tex
\section{Problem formulation}
\label{sec:problem}

We estimate depression severity from the client's language in a counseling transcript. A \emph{turn} is one utterance, and a \emph{session} is one counseling encounter. Single-session estimation uses the current transcript, while longitudinal estimation may also use earlier sessions from the same client. The target is the current session's PHQ-8 score.

\paragraph{Clinical target.}
The eight PHQ-8 symptoms~\citep{kroenke2009phq8} are anhedonia, depressed mood, sleep disturbance, fatigue, appetite disturbance, low self-worth or guilt, concentration difficulty, and psychomotor change. Questionnaire responses are ordinal, with item values in $\{0,1,2,3\}$ and total $Y=\sum_{j=1}^{J}y_j\in[0,24]$, where $J=8$. The synthetic benchmark also uses fractional hidden states and averaged self-reports within these bounds (Appendix~\ref{apdx:longcounsel}).

\paragraph{Prediction task.}
Given the current transcript $\mathcal{T}^{(t)}$ and available history $\mathcal{H}^{(t-1)}=\{\mathcal{T}^{(1)},\ldots,\mathcal{T}^{(t-1)}\}$, we learn
\begin{equation}
f_\theta\bigl(\mathcal{T}^{(t)},\mathcal{H}^{(t-1)}\bigr)\rightarrow
\hat{\mathbf{y}}^{(t)}\in[0,3]^J,\qquad
\hat{Y}^{(t)}=\sum_{j=1}^{J}\hat{y}_j^{(t)}.
\end{equation}
History is empty for the first or only session. EmoTrack uses one previous session by default, with two-session and full-history inputs evaluated in Appendix~\ref{subsec:history-scope}. Past questionnaire labels are not model inputs.

%% file: section/approach.tex
\section{Approach: EmoTrack}
\label{sec:approach}

\begin{figure}[t]
\centering
\includegraphics[width=0.96\linewidth]{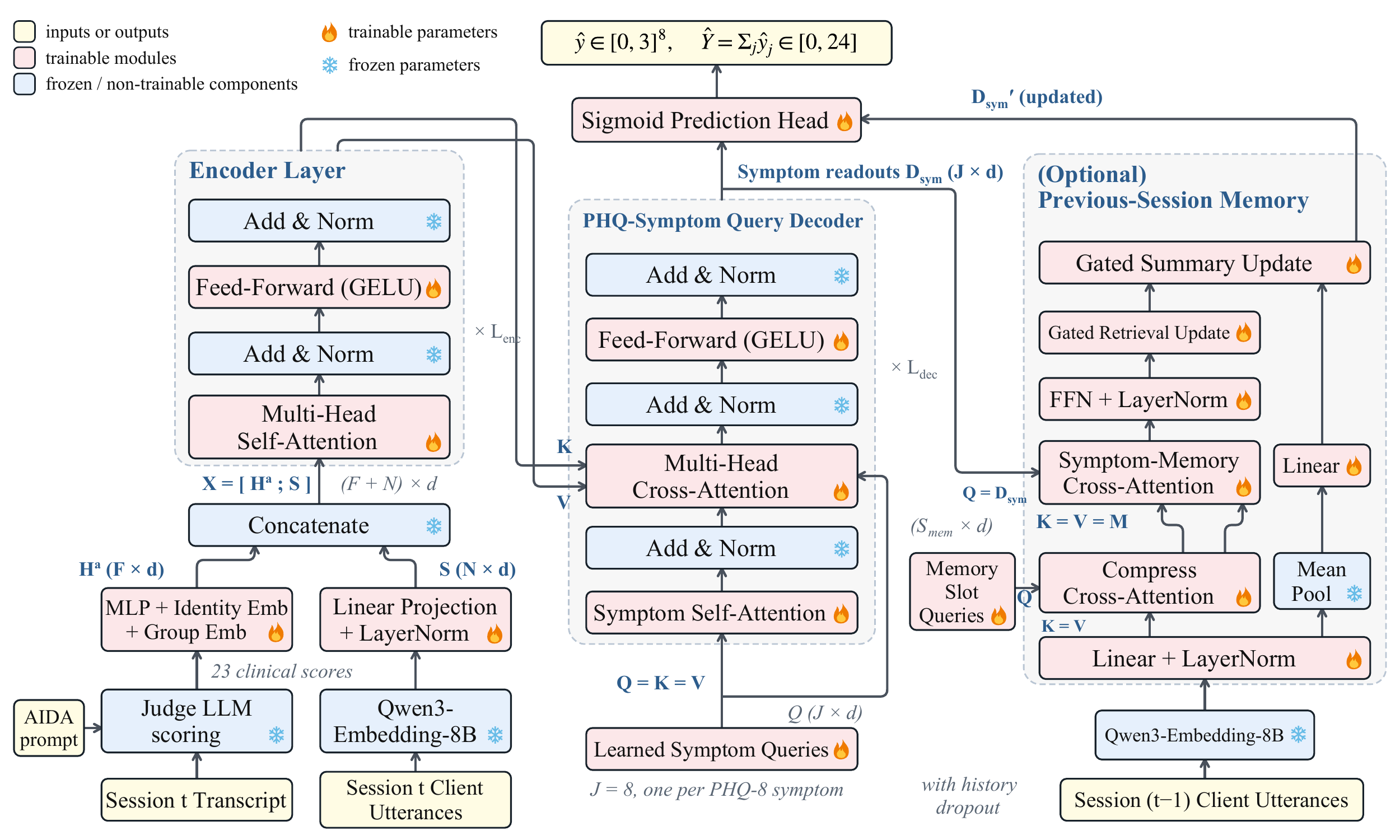}
\caption{EmoTrack architecture for PHQ-8 severity estimation with optional previous-session memory.}
\label{fig:emotrack_overview}
\end{figure}

EmoTrack combines structured clinical features with client-turn semantics to estimate PHQ-8 severity. The two input branches form a shared token sequence, and learned PHQ-8 item queries retrieve symptom-specific evidence from an encoder--decoder backbone. An optional memory module incorporates previous-session context before the prediction head. Figure~\ref{fig:emotrack_overview} summarizes the architecture, and Appendix~\ref{apdx:emotrack-algorithm-detail} details its forward computation.

\subsection{Clinical and semantic inputs}
\label{subsec:input-representation}

The clinical branch provides explicit symptom-related evidence, while the dialogue branch retains semantic information that a fixed feature schema may omit. Both use the current client-side transcript $\mathcal{T}^{(t)}$. The input and speaker ablations in Section~\ref{sec:ablation} test these choices.

\paragraph{Clinical-feature tokens.}
Following AIDA~\citep{lee2026interpretable}, we prompt an LLM to score $F=23$ features comprising eight PHQ-aligned indicators, five linguistic-pattern features, and ten cognitive-distortion markers. Each score lies in $[0,10]$ and is standardized using training-split statistics. For standardized score $z_i$, the token is
\begin{equation}
\mathbf{h}_i=\mathrm{MLP}_{\mathrm{score}}(z_i)+\mathbf{e}^{\mathrm{feat}}_i+
\mathbf{e}^{\mathrm{group}}_{g(i)}.
\end{equation}
The learned feature and group embeddings identify the score and its clinical group $g(i)\in\{1,2,3\}$. Stacking the tokens gives $\mathbf{H}^{\mathrm{c}}\in\mathbb{R}^{F\times d}$.

\paragraph{Client-turn tokens.}
A frozen sentence encoder maps each client turn to $\mathbf{u}_n\in\mathbb{R}^{d_e}$, and a trainable projection gives $\mathbf{s}_n=\mathrm{LayerNorm}(W_{\mathrm{proj}}\mathbf{u}_n+\mathbf{b}_{\mathrm{proj}})$. We remove counselor and empty turns, retain the first $N_{\max}$ client turns in chronological order, and encode each client turn as one fixed-size vector. The resulting sequence is $\mathbf{S}\in\mathbb{R}^{N\times d}$. Validation on \daic selects $N_{\max}=80$, with cap sensitivity shown in Figure~\ref{fig:history-context-ablation}. The longitudinal sessions contain 20 client turns and fit within this cap.

\subsection{Encoder--decoder and PHQ-8 query readout}
\label{subsec:backbone}

Self-attention lets clinical and semantic tokens exchange information before prediction. The encoder processes $\mathbf{X}=[\mathbf{H}^{\mathrm{c}};\mathbf{S}]\in\mathbb{R}^{(F+N)\times d}$ with a padding mask that excludes padded turns from attention. The selected backbone uses two encoder layers (three on \daic) and four decoder layers, with depth ablations reported in Appendix~\ref{subsec:ablation-encoder-decoder-layers}.

Item-specific readout allows each symptom query to retrieve relevant evidence. Eight learned query vectors $\mathbf{Q}\in\mathbb{R}^{J\times d}$, one per PHQ-8 item, cross-attend to the encoded tokens. The decoder returns $\mathbf{D}=[\mathbf{d}_1,\ldots,\mathbf{d}_J]^\top\in\mathbb{R}^{J\times d}$. Table~\ref{tab:ablation-readout} compares this readout with mean pooling, and Table~\ref{tab:aida-qwen-traditional-ml} tests standard regressors on the same upstream inputs.

\subsection{Prediction and training objective}
\label{subsec:training-objective}

After any memory refinement, the prediction head maps each item representation to $\hat{y}_j=3\sigma(\mathbf{w}_j^\top\mathbf{d}_j+b_j)\in[0,3]$, and sums these values into $\hat{Y}=\sum_j\hat{y}_j$. The training objective combines a total-score Huber loss with optional item-level supervision,
\begin{equation}
\label{eq:training-objective}
\mathcal{L}=\underbrace{\mathrm{Huber}(\hat{Y},Y)}_{\mathcal{L}_{\mathrm{agg}}}
+\lambda_{\mathrm{sym}}\underbrace{\frac{1}{J}\sum_{j=1}^{J}\mathrm{Huber}(\hat{y}_j,y_j)}_{\mathcal{L}_{\mathrm{sym}}}.
\end{equation}
The auxiliary term follows symptom-resolved severity estimation~\citep{milintsevich2023towards,agarwal2024analyzing}. Validation selects $\lambda_{\mathrm{sym}}=0.5$ on the main \longcounsel set and $0$ on \daic and the complementary dataset. Thus, the same architecture supports aggregate-only training. We use AdamW~\citep{loshchilov2019decoupled}, learning-rate warmup and decay, gradient clipping, and early stopping on validation loss (Appendix~\ref{sec:add_experiment_setting}).

\paragraph{Optional false-negative penalty.}
When missing a screen-positive session is a separate evaluation concern, training can add
\begin{equation}
\label{eq:fn-aware-loss}
\mathcal{L}_{\mathrm{FN}}=\lambda_{\mathrm{FN}}\mathbf{1}[Y\geq10]\max(0,10-\hat{Y})
\end{equation}
to Eq.~\eqref{eq:training-objective}. The term penalizes a prediction below the PHQ-8 screening threshold of 10 for a reference-positive session~\citep{kroenke2009phq8}. All default comparisons use $\lambda_{\mathrm{FN}}=0$. A separate validation-FNR selection evaluates this option in Section~\ref{subsec:optional-fn}.

\subsection{Optional previous-session memory}
\label{subsec:memory}

For longitudinal inputs, memory refines the current-session item representations using earlier client turns. The default uses the immediately previous session and bypasses memory when history is absent. This keeps the single-session prediction pathway intact. Appendix~\ref{subsec:history-scope} evaluates longer histories with the same compressor.

\paragraph{Compressed retrieval.}
We project previous-session embeddings into $\mathbf{V}^{(t-1)}\in\mathbb{R}^{M\times d}$. Learned slot queries $\mathbf{P}\in\mathbb{R}^{S\times d}$ compress these tokens into fixed-size memory,
\begin{equation}
\mathbf{M}=\mathrm{LN}\bigl(\mathrm{MHA}(\mathbf{P},\mathbf{V}^{(t-1)},\mathbf{V}^{(t-1)})\bigr)\in\mathbb{R}^{S\times d}.
\end{equation}
Here, MHA denotes multi-head attention and LN denotes layer normalization. Current item representations attend to $\mathbf{M}$, producing retrieved evidence $\mathbf{C}\in\mathbb{R}^{J\times d}$. A gated residual update controls its contribution,
\begin{equation}
\mathbf{D}\leftarrow\mathbf{D}+\sigma\bigl(W_g[\mathbf{D};\mathbf{C}]\bigr)\odot W_o\mathbf{C}.
\end{equation}

\paragraph{Summary and regularization.}
A second path mean-pools the projected previous-session tokens into $\bar{\mathbf{v}}\in\mathbb{R}^d$ and broadcasts this vector to all items,
\begin{equation}
\mathbf{D}\leftarrow\mathbf{D}+\sigma\bigl(W_{g'}[\mathbf{D};\bar{\mathbf{v}}]\bigr)\odot\bar{\mathbf{v}}.
\end{equation}
Validation selects the combined retrieval and summary paths, with four slots on \longcounsel and 16 on the complementary dataset. During training, history is dropped with probability $p_{\mathrm{hist}}$ to limit dependence on prior context. The selected rates are $0.1$ and $0.2$, respectively. Figure~\ref{fig:history-context-ablation} shows the dropout sweep. Appendix~\ref{subsec:ablation-memory-mechanism} and Appendix~\ref{subsec:ablation-memory-slots} separate path and capacity effects.

%% file: section/phq_dataset.tex
\section{\longcounsel: dataset construction}
\label{sec:phq-dataset}

We introduce \longcounsel and its construction protocol for controlled current-session PHQ-8 prediction with optional history. Empirical severity trajectories define private symptom states, which guide counseling dialogue and provide a reference for imperfect self-report labels.

\subsection{Construction}
\label{subsec:dataset-construction}

\paragraph{Profiles and symptom trajectories.}
We first convert PsychEval-derived cases~\citep{pan2026psycheval} into stable client profiles and session plans, preserving backgrounds and presenting concerns while rewriting session themes as client-side disclosure beats. We then reconstruct five-visit severity trajectories from PSYCHE-D quarterly observations~\citep{Makhmutova2021PredictingCI}. For each PHQ-9 total-score anchor, we select a complete NHANES DPQ response vector with the matching total~\citep{cdc_nhanes_dpq_l_2024}. Retaining its first eight items produces the hidden PHQ-8 symptom vector, whose sum defines the session's hidden PHQ-8 total. Selecting complete response vectors preserves observed symptom compositions.

\paragraph{Private cues and counseling dialogue.}
Private cues determine symptom content, while RealCBT-derived surface constraints shape how it is expressed. Each symptom score is translated into private behavioral cues describing lived experience without questionnaire wording or numeric labels. We prune candidate cues through joint composition checks and sample five cues per symptom during generation. The client simulator receives these cues, its profile, and cross-session memory. The counselor receives public dialogue context and session guidance. RealCBT~\citep{Wang2025FeelTD} calibrates speaker- and stage-specific turn length, question form, discourse openings, and lexical overlap.

\paragraph{Self-report labels and partitions.}
After each session, the simulated client completes five PHQ-8 passes with shuffled symptom and option orders, and we average the responses itemwise. Models train on these self-reports and are evaluated against the hidden generating state, whose numeric scores are excluded from dialogue. All five visits of a trajectory remain in one data split. Appendix~\ref{apdx:longcounsel-construction} details construction and filtering, and Appendices~\ref{apdx:longcounsel-evaluation}--\ref{apdx:longcounsel-realcbt} report the audits.

%% file: section/dataset_validation.tex
\subsection{Dataset validation}
\label{subsec:dataset-validation}

\begin{figure}[t]
\centering
\subfloat[Self-report error before and after cue pruning.\label{fig:main-cue-error}]{%
  \includegraphics[width=0.32\linewidth]{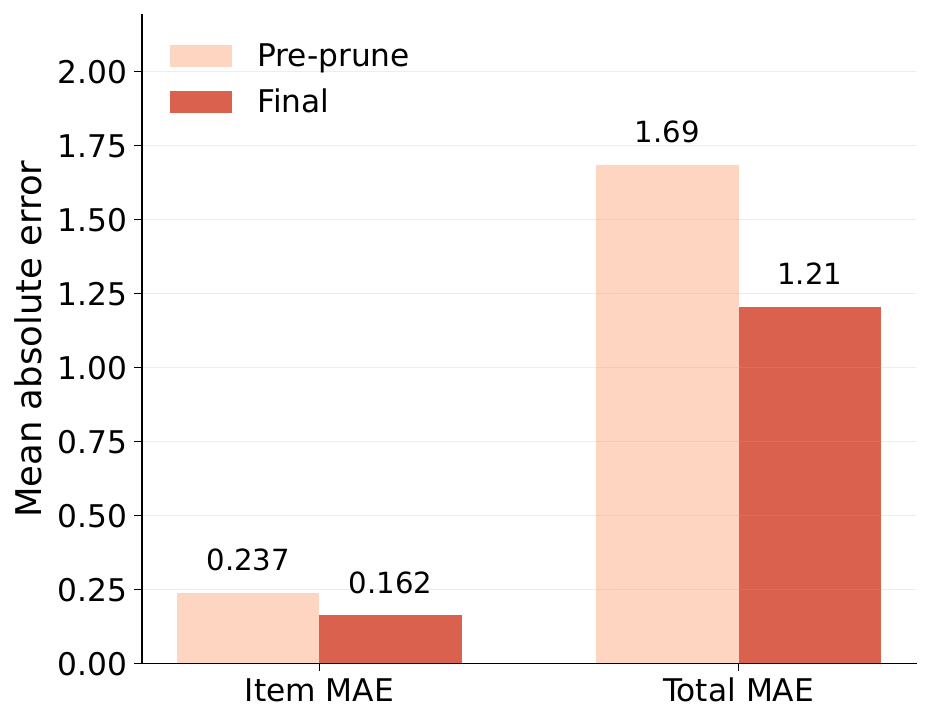}%
}\hfill
\subfloat[Psycholinguistic distance.\label{fig:main-psycholinguistic}]{%
  \includegraphics[width=0.32\linewidth]{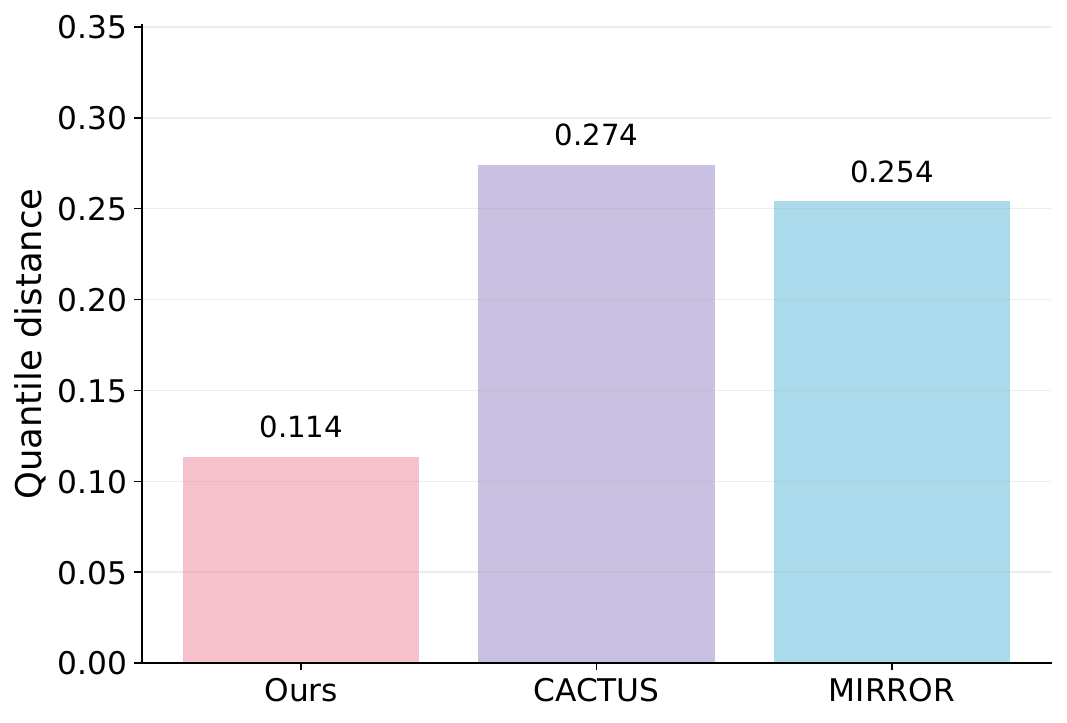}%
}\hfill
\subfloat[Judge distance from chance.\label{fig:main-judge-distance}]{%
  \includegraphics[width=0.32\linewidth]{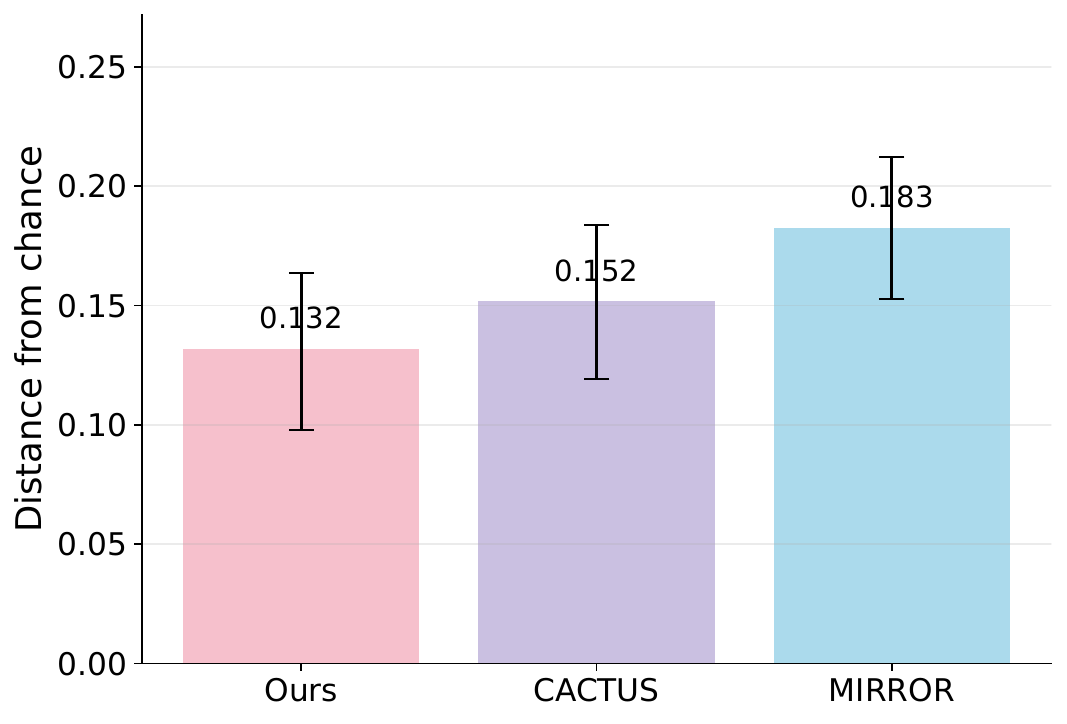}%
}
\caption{Validation of \longcounsel (``Ours'') through self-report fidelity and dialogue similarity to RealCBT. Lower values indicate smaller errors or closer alignment with RealCBT.}
\label{fig:dataset-validation}
\end{figure}

The longitudinal benchmark requires reliable simulated labels and dialogue that resembles real counseling. Figure~\ref{fig:dataset-validation} evaluates these properties through three complementary audits, with the full analysis in Appendices~\ref{apdx:longcounsel-evaluation}--\ref{apdx:longcounsel-realcbt}.

\paragraph{Self-report fidelity.}
Cue pruning substantially improves agreement between simulated self-reports and prescribed symptom states. Figure~\ref{fig:main-cue-error} reports item-level and total-score MAE between simulated self-reports and the hidden PHQ-8 vector when sampling five behavioral cues per symptom. Pruning reduces item MAE by $31.6\%$ and total-score MAE by $28.4\%$ while consistently increasing exact-match rates across item, profile, and total-score levels. Appendix Figure~\ref{fig:cue-composition} provides the full exact-match and per-symptom breakdown.

\paragraph{Dialogue realism.}
\longcounsel aligns more closely with real counseling transcripts in RealCBT than CACTUS and MIRROR across both linguistic and judge-based realism measures~\citep{Wang2025FeelTD,lee2024cactus,kim2025mirror}. Figure~\ref{fig:main-psycholinguistic} measures aggregate psycholinguistic quantile distance over speaker-specific turn length, readability, and affective statistics, while Figure~\ref{fig:main-judge-distance} reports pointwise judge distance from chance to account for systematic judgment inversion. \longcounsel more than halves the psycholinguistic distance relative to either baseline and reduces judge distance from chance by $13.2\%$ relative to CACTUS and $27.9\%$ relative to MIRROR. Appendix Figure~\ref{fig:feature-distance} details the feature-level audit across all seven psycholinguistic statistics.

%% file: section/experiment.tex
\section{Experiments}
\label{sec:experiment}

We first outline the experimental setup, including the five benchmark datasets, evaluation protocols, and baseline methods (Section~\ref{subsec:experiment-settings}). We then evaluate EmoTrack's main prediction accuracy across real single-session interviews, zero-shot protocol transfer, and multi-session longitudinal counseling (Section~\ref{subsec:performance-comparison}), followed by an analysis of the optional false-negative penalty for clinical screening (Section~\ref{subsec:optional-fn}) and systematic component ablations (Section~\ref{sec:ablation}). Extended evaluations on per-symptom bias (Appendix~\ref{subsec:symptom-bias}), session-to-session transitions (Appendix~\ref{subsec:transitions}), and trajectory stratification (Appendix~\ref{subsec:trajectory-strat}) are deferred to the appendices.

\subsection{Experimental settings}
\label{subsec:experiment-settings}

\paragraph{Datasets and protocols.}
We evaluate on five datasets. \daic~\citep{Gratch2014TheDA} provides 189 real PHQ-8-labeled interviews, with 107/35/47 training/validation/test participants. For \textsc{E-DAIC}~\citep{ringeval2019avec}, models trained and selected on \daic are evaluated without adaptation. For multi-session evaluation, we use \longcounsel, generated with Qwen3.5-35B-A3B. LongCounsel-8~\citep{li2026longcounsel} extends \longcounsel with additional dialogue generators. We use two \href{https://huggingface.co/datasets/hiddensev/LongCounsel-8}{datasets} from this work, LongCounsel-8-Luna and LongCounsel-8-GPT54-Mini (the complementary set), for cross-generator evaluation. \longcounsel and LongCounsel-8-Luna each contain 3,690 trajectories (18,450 sessions), with the latter generated using GPT-5.6 Luna. LongCounsel-8-GPT54-Mini uses GPT-5.4-mini and contains 369 trajectories (1,845 sessions). All use 70/10/20\% trajectory-level splits, with five sessions per trajectory and 20 client plus 20 counselor turns per session. Models train on simulated self-reports and are scored against hidden generating states. History contains earlier transcripts without past labels. The GPT-5.4-mini set differs in generator and scale.

\paragraph{Model inputs and training.}
The frozen Qwen3-Embedding-8B encoder~\citep{zhang2025qwen3embedding} produces a 4,096-dimensional vector per client turn. Qwen3.5-35B-A3B~\citep{qwen3.5} extracts the 23 AIDA clinical scores. Each extraction seed is paired across downstream methods. We train with AdamW, batch size 32, learning rate $10^{-3}$, weight decay $10^{-2}$, 5\% warmup followed by linear decay, gradient clipping at norm 1.0, and early stopping with patience 8 on validation loss. Appendix~\ref{sec:add_experiment_setting} gives dataset-specific configurations and hardware.

\paragraph{Metrics and selection.}
We report current-session PHQ-8 test MAE and screening rates as five-seed means, with error bars and $\pm$ values denoting one sample standard deviation across seeds. Transition and trajectory-type analyses stratify current-session score error. We select hyperparameters by validation MAE and apply early stopping on validation loss. Normalized aggregate error summarizes total-score accuracy across the five sets,
\begin{equation}
\operatorname{NAE}=\sqrt{\frac{1}{5}\sum_{k=1}^{5}\left(\frac{\mathrm{MAE}_k}{U_k}\right)^2},
\end{equation}
where $U_k$ is the test MAE of a constant predictor that outputs the training-label median for set $k$. \textsc{E-DAIC} transfer retains the \daic training median. Each set has equal weight, lower is better, and one denotes the constant predictor. The optional screening analysis also reports false-negative rate (FNR) among reference-positive sessions and false-positive rate (FPR) among reference-negative sessions at a PHQ-8 threshold of 10. Its penalty weight is selected separately by validation FNR.

\paragraph{Baselines.}
For the clinical-based baselines, \textbf{AIDA}~\citep{lee2026interpretable} fits ridge regression to 23 extracted features, and \textbf{LMIQ}~\citep{rosenman2024llm} fits a random forest to LLM-completed questionnaire features. For the training-based baselines, \textbf{Lau et al.}~\citep{lau2023automatic} use a tuned RoBERTa question--response encoder and sentence-transformer branch. \textbf{Milintsevich et al.}~\citep{milintsevich2023towards} use hierarchical symptom prediction. \textbf{EnsemBERT}~\citep{ravenda2025transforming} combines symptom-relevant retrieval with attention, adapted here from BDI-II to PHQ-8. For AIDA and LMIQ, ``now'' uses the current session and ``all'' concatenates available history. Where applicable, trained baselines use both original encoders and Qwen3 embeddings. Comparisons share splits and extraction seeds.

\subsection{Main results}
\label{subsec:performance-comparison}

\begin{figure}[t]
\centering
\subfloat[Cross-setting MAE.\label{fig:cross-regime-radar}]{%
  \includegraphics[width=0.65\textwidth]{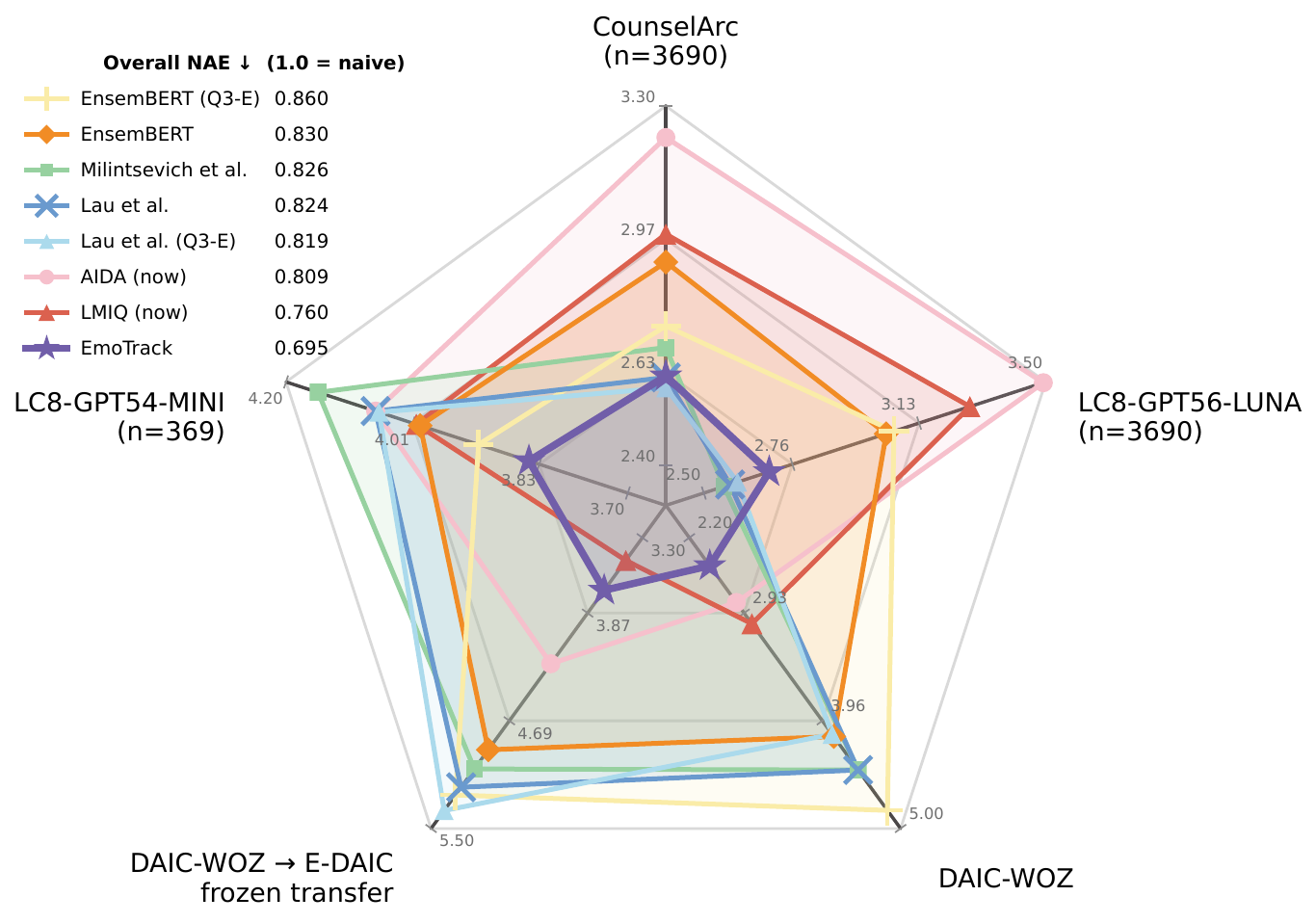}%
}\hfill
\begin{minipage}[b]{0.31\textwidth}
  \centering
  \subfloat[Input source.\label{fig:ablation-input-source}]{%
  \includegraphics[width=\linewidth]{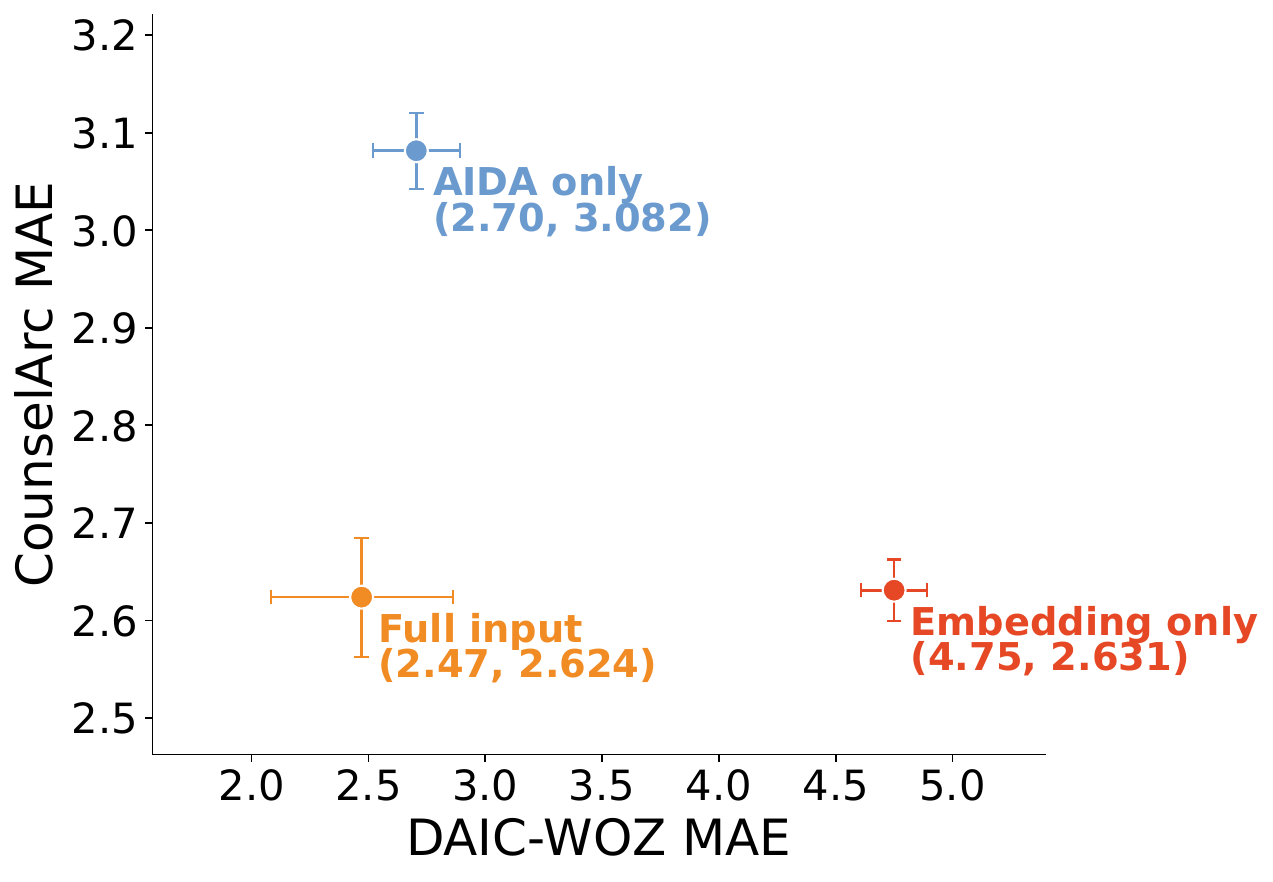}%
}
  \par
  \subfloat[Speaker selection.\label{fig:ablation-speaker-view}]{%
  \includegraphics[width=\linewidth]{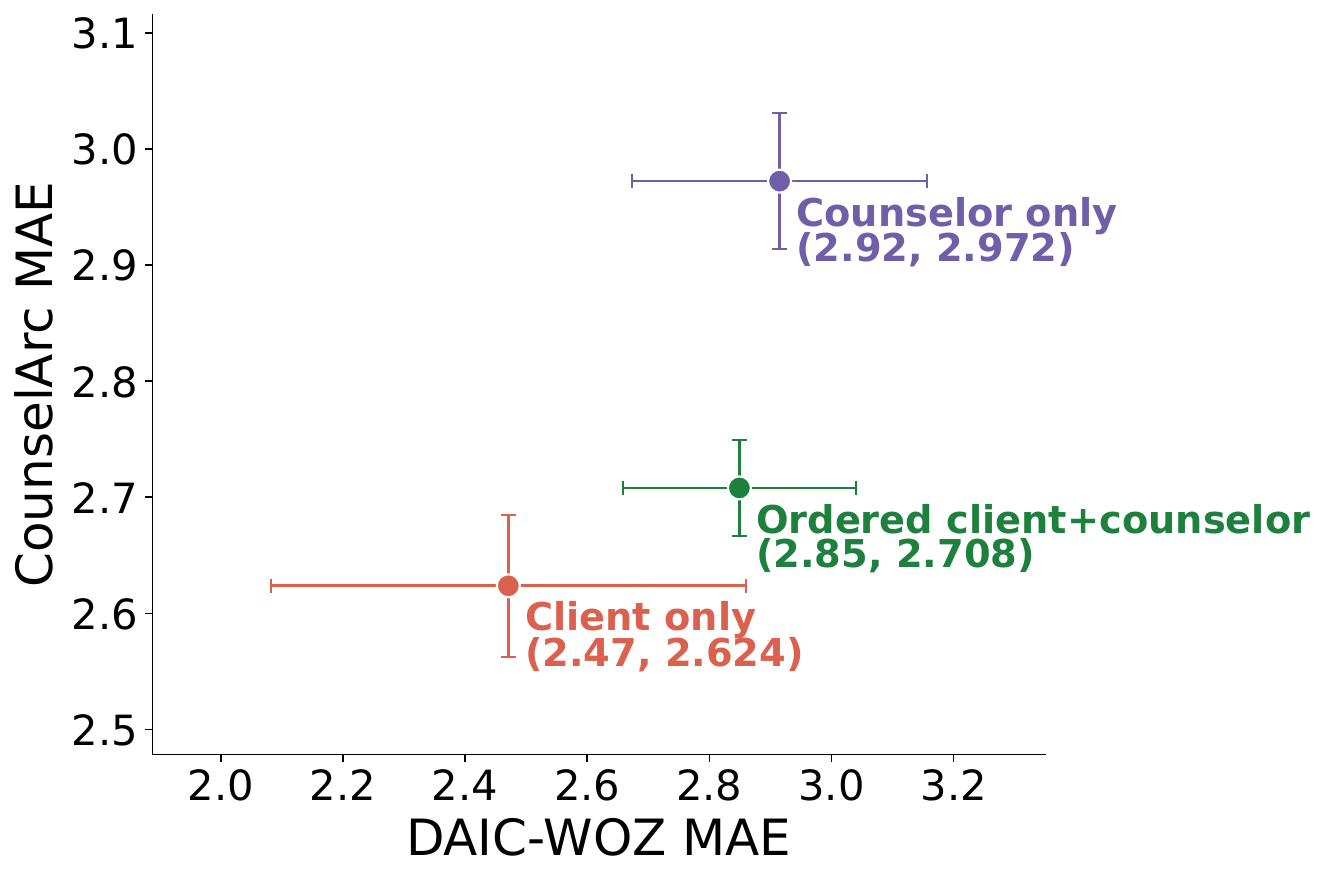}%
}
\end{minipage}
\caption{Total-score MAE across evaluation sets (a), input sources (b), and speaker selections (c). Point labels in (b,c) give $(\text{\daic MAE},\text{\longcounsel MAE})$.}
\label{fig:ablation-and-overall-results}
\end{figure}

\paragraph{Accuracy across settings.}
EmoTrack generalizes more effectively across single-session, transfer, and longitudinal settings than either clinical-based or training-based baselines alone. Figure~\ref{fig:cross-regime-radar} presents the cross-setting MAE alongside NAE, which measures overall cross-dataset accuracy. Whereas baselines struggle outside their primary regime, EmoTrack encloses the smallest radar footprint and achieves the lowest NAE, outperforming the strongest overall competitor, LMIQ (now), by approximately $8.6\%$. Appendix~\ref{subsec:all-results} details the full per-dataset results.

\paragraph{Real single-session evaluation.}
Jointly modeling clinical features and client-turn semantics yields the most accurate severity predictions on real clinical interviews. Table~\ref{tab:baseline-comparison-daic} reports the five-seed test MAE on \daic, evaluating single-session estimation under scarce supervision. Whereas purely training-based predictors lag behind prompted clinical methods, EmoTrack achieves the lowest mean error and reduces MAE by $12.5\%$ relative to the strongest baseline, AIDA (now).

\input{table/daic_woz_mae_comparison}

\paragraph{Transfer without adaptation.}
EmoTrack generalizes competitively to unseen interview protocols without fine-tuning, ranking second overall and substantially outperforming all training-based baselines. Figure~\ref{fig:cross-regime-radar} shows the frozen \daic-to-\textsc{E-DAIC} transfer MAE, measuring out-of-distribution accuracy when models trained on \daic are evaluated directly on \textsc{E-DAIC}. EmoTrack ranks second overall, trailing only LMIQ (now) by $6.5\%$ while reducing MAE by $13.1\%$ relative to AIDA (now) and by at least $24.6\%$ relative to every training-based baseline. Appendix~\ref{subsec:all-results} details the full results.

\paragraph{Longitudinal evaluation.}
EmoTrack maintains strong performance across diverse synthetic dialogue generators and achieves the lowest error under reduced supervision. Figure~\ref{fig:cross-regime-radar} displays the multi-session test MAE across the three longitudinal benchmarks, evaluating current-session estimation with prior-session context. On the smaller complementary set (LongCounsel-8-GPT54-Mini), EmoTrack ranks first and reduces MAE by $1.9\%$ relative to the strongest baseline, EnsemBERT (Qwen3-Emb), while staying competitive to the leading trained predictors on the two full-scale corpora. Appendix~\ref{subsec:all-results} reports the complete longitudinal comparison.

\subsection{Optional false-negative penalty}
\label{subsec:optional-fn}

\begin{figure}[t]
\centering
\includegraphics[width=\linewidth]{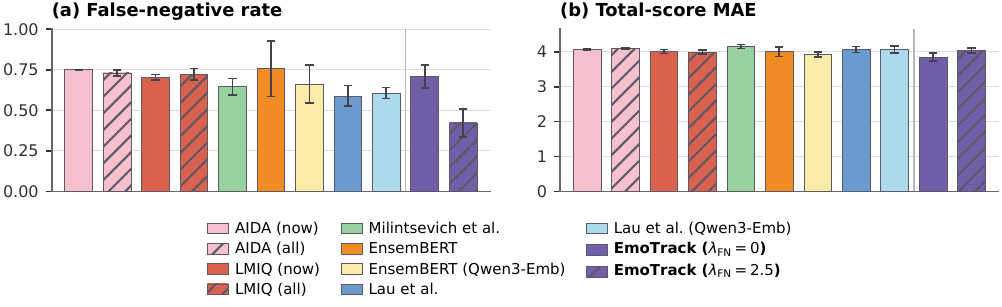}
\caption{False-negative rate (a) and total-score MAE (b) on the complementary set for the baselines and EmoTrack with and without the optional penalty.}
\label{fig:optional-fn}
\end{figure}

Applying the optional false-negative penalty markedly improves screening sensitivity for reference-positive sessions while introducing a controlled trade-off in score error. Figure~\ref{fig:optional-fn} presents the false-negative rate (FNR) and total-score MAE across all methods on the complementary set. Relative to the default configuration ($\lambda_{\mathrm{FN}}=0$), validation-selected $\lambda_{\mathrm{FN}}=2.5$ lowers FNR by $40.4\%$—surpassing all baselines including the strongest competitor, Lau et al.—while increasing MAE by $5.0\%$. Appendix Table~\ref{tab:fn-aware-comparison} reports the accompanying FPR trade-off, and Appendix~\ref{subsec:fn-sweep} provides the weight-sensitivity analysis.

\subsection{Ablation study}
\label{sec:ablation}

In this section, we ablate input sources, speaker selection, predictor architecture, history dropout, retained client turns, and PHQ-symptom query decoding. Further ablations covering encoder--decoder depth (Appendix~\ref{subsec:ablation-encoder-decoder-layers}), previous-session memory pathways (Appendix~\ref{subsec:ablation-memory-mechanism}), memory slot capacity (Appendix~\ref{subsec:ablation-memory-slots}), historical session scope (Appendix~\ref{subsec:history-scope}), symptom-wise loss weighting (Appendix~\ref{subsec:ablation-item-loss}), clinical-feature extractors (Appendix~\ref{subsec:extractor-swap}), and inference runtime (Appendix~\ref{subsec:runtime-compute-cost}) are reported in the appendices.

\paragraph{Input sources and speaker selection.}
The relative contribution of clinical scores and client-turn embeddings shifts with supervision scale, whereas client-only speech is consistently optimal across regimes. Figure~\ref{fig:ablation-input-source} compares test MAE on \daic and \longcounsel when ablating each input branch, and Figure~\ref{fig:ablation-speaker-view} evaluates alternative speaker selections. On \daic, removing clinical features nearly doubles MAE while removing turn embeddings has a modest effect; conversely, on \longcounsel, turn embeddings alone nearly match the full model whereas clinical features alone perform worse. For speaker selection, using only client turns achieves the lowest mean error on both datasets.

\paragraph{Predictor architecture.}
EmoTrack's backbone extracts more predictive signal from the combined inputs than standard tabular and neural regressors. Table~\ref{tab:aida-qwen-traditional-ml} compares overall test MAE on \longcounsel and \daic when EmoTrack, ridge regression, an MLP, and gradient-boosted trees are trained on identical clinical scores and client-turn embeddings. EmoTrack outperforms all three baselines, reducing MAE by $5.9\%$ on \longcounsel and $12.0\%$ on \daic relative to gradient-boosted trees, the strongest standard regressor.

\input{table/ablations/aida_qwen_traditional_ml_seed0_4}

\input{section/main_ablation_figures}

\paragraph{History dropout and retained client turns.}
Moderate history dropout and bounded client-turn context improve generalization by limiting reliance on noisy or excessive dialogue history. Figure~\ref{fig:history-context-ablation} plots total-score test MAE across history-dropout rates on the longitudinal sets and retained-client-turn caps on \daic. A dropout rate of $0.1$ achieves the lowest error on \longcounsel, reducing MAE by $2.2\%$ relative to no dropout, whereas the complementary set reaches its test minimum without dropout and selects $0.2$ by validation ($1.4\%$ higher). On \daic, retaining 80 client turns outperforms both shorter and longer contexts, reducing MAE by $4.6\%$ relative to 384 client turns and by at least $9.3\%$ relative to 20 or 40 client turns while accommodating all 20-turn longitudinal sessions. Appendices~\ref{subsec:ablation-history-dropout} and~\ref{subsec:ablation-max-context-sentences} detail these sweeps.

\paragraph{PHQ-symptom query decoder.}
Decoding symptom-specific representations through learned PHQ-8 queries outperforms uniform mean pooling on the primary single-session and longitudinal benchmarks. Table~\ref{tab:ablation-readout} compares total-score test MAE between the PHQ-symptom query decoder and a mean-pooled encoder head across three datasets. Replacing the query decoder with mean pooling increases MAE by $4.3\%$ on \longcounsel and $8.8\%$ on \daic, while the two readouts differ by only $0.5\%$ on the complementary set.

\input{table/ablations/table_readout}

%% file: table/daic_woz_mae_comparison.tex
\begin{table}[ht]
\centering
\small
\caption{Total-score MAE on \daic, with the lowest mean in bold. Values are mean $\pm$ sample standard deviation over five seeds.}
\label{tab:baseline-comparison-daic}
\setlength{\tabcolsep}{1.9pt}
\begin{tabular}{@{}lcccccccc@{}}
\toprule
Metric
& \makecell{AIDA\\(now)}
& \makecell{LMIQ\\(now)}
& \makecell{Milintsevich\\et al.}
& EnsemBERT
& \makecell{EnsemBERT\\(Qwen3-Emb)}
& \makecell{Lau\\et al.}
& \makecell{Lau et al.\\(Qwen3-Emb)}
& \makecell{\textbf{EmoTrack}\\\textbf{(Ours)}} \\
\midrule
MAE $\downarrow$
& \makecell{2.8234\\{\small $\pm$ 0.2099}}
& \makecell{3.0285\\{\small $\pm$ 0.0460}}
& \makecell{4.4370\\{\small $\pm$ 0.1197}}
& \makecell{4.1149\\{\small $\pm$ 0.3339}}
& \makecell{4.8255\\{\small $\pm$ 0.7277}}
& \makecell{4.4367\\{\small $\pm$ 0.5744}}
& \makecell{4.0927\\{\small $\pm$ 0.2795}}
& \makecell{\textbf{2.4708}\\{\small \textbf{$\pm$ 0.3892}}} \\
\bottomrule
\end{tabular}
\end{table}

%% file: table/ablations/aida_qwen_traditional_ml_seed0_4.tex
\begin{table}[ht]
\centering
\small
\caption{Predictor comparison using the same AIDA features and Qwen3 embeddings on \longcounsel and \daic.}
\label{tab:aida-qwen-traditional-ml}
\begin{tabular}{lcc}
\toprule
\multirow{2}{*}{\textbf{Method}} & \multicolumn{2}{c}{\textbf{Overall MAE $\downarrow$}} \\
\cmidrule(l){2-3}
& \longcounsel & \daic \\
\midrule
Ridge regression & 3.1709 $\pm$ 0.0027 & 3.1806 $\pm$ 0.0119 \\
MLP regressor & 2.9677 $\pm$ 0.0262 & 3.8103 $\pm$ 0.2072 \\
Gradient-boosted trees & 2.7881 $\pm$ 0.0092 & 2.8070 $\pm$ 0.1136 \\
\textbf{EmoTrack} & \textbf{2.6238 $\pm$ 0.0610} & \textbf{2.4708 $\pm$ 0.3892} \\
\bottomrule
\end{tabular}
\end{table}

%% file: section/main_ablation_figures.tex
\begin{figure}[t]
  \centering
  \includegraphics[width=\linewidth]{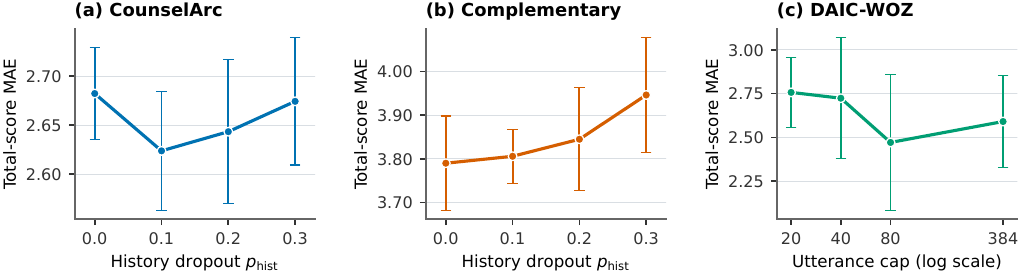}
  \caption{Total-score MAE by history-dropout rate on \longcounsel (a) and the complementary set (b), and by retained-client-turn cap on \daic (c). Panel (c) uses a logarithmic horizontal axis.}
  \label{fig:history-context-ablation}
\end{figure}

%% file: table/ablations/table_readout.tex
\begin{table}[ht]
\centering
\small
\caption{
Effect of replacing the PHQ-symptom query decoder with a mean-pooled encoder head on \longcounsel, \daic, and the complementary GPT-5.4-mini-generated dataset.
}
\label{tab:ablation-readout}
\begin{tabular}{lccc}
\toprule
\multirow{2}{*}{\textbf{Prediction head}} & \multicolumn{3}{c}{\textbf{Overall MAE $\downarrow$}} \\
\cmidrule(l){2-4}
& \longcounsel & \daic & \textsc{GPT-5.4-mini Comp.} \\
\midrule
PHQ-symptom query decoder & \textbf{2.6238 $\pm$ 0.0610}~\textsuperscript{\ensuremath{\dagger}} & \textbf{2.4708 $\pm$ 0.3892}~\textsuperscript{\ensuremath{\dagger}} & 3.8447 $\pm$ 0.1181~\textsuperscript{\ensuremath{\dagger}} \\
Mean-pooled encoder head & 2.7364 $\pm$ 0.0411 & 2.6889 $\pm$ 0.1328 & \textbf{3.8262 $\pm$ 0.1046} \\
\bottomrule
\end{tabular}
\end{table}

%% file: section/conclusion.tex
\section{Conclusion}
\label{sec:conclusion}

EmoTrack bridges prompted clinical structure and client-turn semantics through identity-preserving joint encoding and PHQ-8 item queries, with an optional gated memory module for longitudinal use. We also introduce the \longcounsel benchmark and its construction protocol for controlled session-level severity prediction, supported by comprehensive cue-fidelity and dialogue-similarity audits. Evaluation across five diverse datasets consistently yields the lowest normalized aggregate error, while ablations highlight dataset-dependent gains from historical memory and supervision.

%% file: section/add_related_work.tex
\section{Additional related work}
\label{sec:add_related}

\subsection{Transcript-based depression severity prediction}
\label{subsec:add_text_score_methods}

Table~\ref{tab:lit-emotion-tracking} compares transcript-based score prediction with multimodal and classification approaches, identifying the input evidence and prediction target for each method.

LMIQ fits a random forest to prompted questionnaire responses~\citep{rosenman2024llm}, and AIDA fits a regression model to prompted clinical-feature scores~\citep{lee2026interpretable}. PsychiatricPrism trains a transformer and evaluates symptom predictions against expert annotations~\citep{agarwal2024analyzing}. EmoTrack combines clinically structured features with supervised symptom prediction and retains client turns as separate evidence tokens.

Interpretation of transcript results also depends on how the interview was conducted. Burdisso et al.~\citep{burdisso2024daic} identify interviewer-prompt shortcuts in \daic depression classification, motivating explicit checks of speaker selection. Milintsevich et al.~\citep{milintsevich2023towards} further observe that some symptoms are sparsely expressed in the transcript and that prediction errors vary by symptom. These findings motivate our speaker-view and per-symptom analyses.

\input{table/depression_track_studies}

\subsection{Counseling and motivational interviewing datasets}
\label{subsec:add_counseling_datasets}

Table~\ref{tab:counseling-datasets} surveys counseling and motivational-interviewing resources by provenance, language, multi-session structure, and PHQ supervision. Resources based on human material include public demonstrations~\citep{Malhotra2021SpeakerAT,Srivastava2022CounselingSU,Wang2025FeelTD}, trained-counselor role-play~\citep{Qi2025KokoroChatAJ,Wu2022AnnoMIAD,Gunal2025ExaminingSC}, reconstructed or rewritten conversations~\citep{Liu2023ChatCounselorAL,qiu2025psydial}, and licensed therapy transcripts~\citep{CPTFullCollection}. The \emph{real} provenance category includes role-play and reconstructed source material. Synthetic resources use different therapeutic frameworks and generation procedures~\citep{lee2024cactus,zhang2024cpsycoun,kim2025mirror,abbasi2025hamraz,xie2025psydt,kim2025kmi,bn2025real,suhas2025thousand,pan2026psycheval}.

These corpora support research on dialogue quality, therapeutic style, and conversational continuity. Within the resources summarized here, \daic~\citep{Gratch2014TheDA} provides real single-session PHQ supervision, while \longcounsel supplies synthetic multi-session targets prescribed during construction (Section~\ref{sec:phq-dataset}). Section~\ref{subsec:experiment-settings} specifies the resource versions used in our experiments.

\input{table/counseling_dataset_studies.tex}

%% file: table/depression_track_studies.tex
\begin{table}[ht]
  \centering
  \small
  \caption{Approaches to depression evaluation.}
  \label{tab:lit-emotion-tracking}
  \setlength{\tabcolsep}{5pt}
  \begin{tabular}{lllll}
    \toprule
    \textbf{Study} &
    \textbf{Modality} &
    \textbf{Data} &
    \textbf{Clinical Criteria} &
    \textbf{Output}
    \\
    \midrule

    \multicolumn{5}{l}{\textit{Multimodal methods}} \\
    \midrule

    \makecell[l]{MoodCapture~\cite{nepal2024moodcapture}} &
    Image &
    Smartphone images &
    PHQ-8, DSM-5 &
    Score \\

    \makecell[l]{LLM+Landmarks~\cite{zhang2024llms}} &
    Speech + Text &
    \daic &
    PHQ-8 &
    Binary \\

    \makecell[l]{Sadeghi et al.~\cite{sadeghi2024harnessing}} &
    Video + Speech + Text &
    E-DAIC &
    PHQ-8 &
    Score \\

    \makecell[l]{SEGA~\cite{Chen2024DepressionDI}} &
    Video + Speech + Text &
    \daic &
    PHQ-8 &
    Score \\
    
    \midrule
    \multicolumn{5}{l}{\textit{Text-based methods with classification output}} \\
    \midrule

    \makecell[l]{DORIS~\cite{lan2025depression}} &
    Text &
    Weibo \& Twitter posts &
    DSM-5 &
    Binary \\

    \makecell[l]{Wang et al.~\cite{Wang2024ExplainableDD}} &
    Text &
    Social-media posts &
    BDI &
    Binary \\

    \makecell[l]{RED~\cite{zhang2025explainable}} &
    Text &
    \daic &
    PHQ-8 &
    Binary \\

    \makecell[l]{PsyGUARD~\cite{qiu2024psyguard}} &
    Text &
    Counseling utterances &
    C-SSRS &
    Level \\

    \midrule
    \multicolumn{5}{l}{\textit{Text-based methods with score output}} \\
    \midrule

    \makecell[l]{Lau et al.~\cite{lau2023automatic}} &
    Text &
    \daic &
    PHQ-8 &
    Score \\

    \makecell[l]{PsychiatricPrism~\cite{agarwal2024analyzing}} &
    Text &
    \daic &
    PHQ-8 &
    Score \\

    \makecell[l]{LMIQ~\cite{rosenman2024llm}} &
    Text &
    \daic &
    PHQ-8 &
    Score \\

    \makecell[l]{AIDA~\cite{lee2026interpretable}} &
    Text &
    \daic &
    PHQ-8 &
    Score \\

    \makecell[l]{Milintsevich et al. ~\cite{milintsevich2023towards}} &
    Text &
    \daic &
    PHQ-8 &
    Score \\
    
    \makecell[l]{EnsemBERT~\cite{ravenda2025transforming}} &
    Text &
    Reddit posts &
    BDI-II &
    Score \\

    \midrule
    EmoTrack (ours) & Text & \makecell[l]{\daic \&\\ \longcounsel} & PHQ-8 & Score \\

    \bottomrule
  \end{tabular}
\end{table}

%% file: table/counseling_dataset_studies.tex
\begin{table}[ht]
  \centering
  \small
  \caption{Counseling and motivational interviewing dialogue resources.}
  \label{tab:counseling-datasets}
  \begin{tabular}{l l l cc}
    \toprule
    \textbf{Work} &
    \textbf{Type} &
    \textbf{Lang.} &
    \textbf{Multi-session} &
    \textbf{PHQ-supervised}
    \\
    \midrule

    \multicolumn{5}{l}{\textit{Real counseling / motivational interviewing resources}} \\
    \midrule

    \makecell[l]{KokoroChat~\cite{Qi2025KokoroChatAJ}} &
    Real &
    JA &
    \cmark &
    \xmark \\

    \makecell[l]{HOPE~\cite{Malhotra2021SpeakerAT}} &
    Real &
    EN &
    \xmark &
    \xmark \\

    \makecell[l]{MEMO~\cite{Srivastava2022CounselingSU}} &
    Real &
    EN &
    \xmark &
    \xmark \\

    \makecell[l]{Psych8K~\cite{Liu2023ChatCounselorAL}} &
    Real &
    EN &
    \xmark &
    \xmark \\

    \makecell[l]{PsyDial-D0\_m~\cite{qiu2025psydial}} &
    Real &
    ZH &
    \cmark &
    \xmark \\

    \makecell[l]{Counseling and\\Psychotherapy\\Transcripts~\cite{CPTFullCollection}} &
    Real &
    EN &
    \cmark &
    \xmark \\

    \makecell[l]{RealCBT~\cite{Wang2025FeelTD}} &
    Real &
    EN &
    \xmark &
    \xmark \\

    \makecell[l]{AnnoMI~\cite{Wu2022AnnoMIAD}} &
    Real &
    EN &
    \xmark &
    \xmark \\

    \makecell[l]{MIDAS~\cite{Gunal2025ExaminingSC}} &
    Real &
    ES &
    \xmark &
    \xmark \\

    \makecell[l]{MentalChat16K~\cite{Xu2025MentalChat16KAB}} &
    Mixed &
    EN &
    \xmark &
    \xmark \\

    \makecell[l]{\daic~\cite{Gratch2014TheDA}} &
    Real &
    EN &
    \xmark &
    \cmark \\

    \midrule
    \multicolumn{5}{l}{\textit{Synthetic counseling / motivational interviewing resources}} \\
    \midrule

    \makecell[l]{PsyDT / PsyDTCorpus~\cite{xie2025psydt}} &
    Synthetic &
    ZH &
    \xmark &
    \xmark \\

    \makecell[l]{PsyDial~\cite{qiu2025psydial}} &
    Synthetic &
    ZH &
    \cmark &
    \xmark \\

    \makecell[l]{CACTUS~\cite{lee2024cactus}} &
    Synthetic &
    EN &
    \xmark &
    \xmark \\

    \makecell[l]{CPsyCoun /\\Memo2Demo~\cite{zhang2024cpsycoun}} &
    Synthetic &
    ZH &
    \xmark &
    \xmark \\

    \makecell[l]{HamRaz~\cite{abbasi2025hamraz}} &
    Synthetic &
    FA &
    \xmark &
    \xmark \\

    \makecell[l]{MIRROR~\cite{kim2025mirror}} &
    Synthetic &
    EN &
    \xmark &
    \xmark \\

    \makecell[l]{KMI~\cite{kim2025kmi}} &
    Synthetic &
    KO &
    \xmark &
    \xmark \\

    \makecell[l]{Thousand Voices of\\Trauma~\cite{suhas2025thousand}} &
    Synthetic &
    EN &
    \xmark &
    \xmark \\

    \makecell[l]{PsychEval~\cite{pan2026psycheval}} &
    Synthetic &
    ZH &
    \cmark &
    \xmark \\

    \midrule
    \longcounsel (ours) &
    Synthetic &
    EN &
    \cmark &
    \cmark \\

    \bottomrule
  \end{tabular}
\end{table}

%% file: section/add_algorithm.tex
\section{Algorithmic details of EmoTrack}
\label{apdx:emotrack-algorithm-detail}

Algorithm~\ref{alg:emotrack} summarizes the forward computation using the tokenizers, backbone, and memory module defined in Section~\ref{sec:approach}.

\begin{algorithm}[ht]
\caption{EmoTrack forward computation}
\label{alg:emotrack}
\begin{algorithmic}[1]
\REQUIRE Current transcript $\mathcal{T}^{(t)}$; optional previous transcript $\mathcal{T}^{(t-1)}$; learned PHQ-8 symptom queries $\mathbf{Q}\in\mathbb{R}^{J\times d}$
\ENSURE PHQ-8 symptom predictions $\hat{\mathbf{y}}\in[0,3]^J$, where $J=8$

\STATE \textbf{// Current-session representation}
\STATE $\mathbf{a}^{(t)} \gets \phi_{\mathrm{LLM}}(\mathcal{T}^{(t)})$
\STATE $\mathbf{H}^{\mathrm{c}} \gets \Phi_{\mathrm{clin}}(\mathbf{a}^{(t)})$
\STATE $\mathbf{S}^{(t)} \gets \Phi_{\mathrm{dlg}}\!\left(E(\mathcal{T}^{(t)})\right)$
\STATE $\mathbf{X}^{(t)} \gets [\mathbf{H}^{\mathrm{c}};\mathbf{S}^{(t)}]$

\STATE \textbf{// Current-session PHQ-8 symptom encoding}
\STATE $\mathbf{D}^{(t)} \gets \mathcal{B}_{\theta}(\mathbf{X}^{(t)},\mathbf{Q})$

\STATE \textbf{// Optional previous-session memory refinement}
\IF{$\mathcal{T}^{(t-1)}$ is available and not dropped by history dropout}
    \STATE $\mathbf{V}^{(t-1)} \gets \Phi_{\mathrm{hist}}\!\left(E(\mathcal{T}^{(t-1)})\right)$
    \STATE $\mathbf{D}^{(t)} \gets \mathcal{R}_{\theta}\!\left(\mathbf{D}^{(t)},\mathbf{V}^{(t-1)}\right)$
\ENDIF

\STATE \textbf{// Symptom-level PHQ-8 prediction}
\STATE $\hat{y}_j \gets 3\cdot\sigma(\mathbf{w}_j^\top \mathbf{d}_j^{(t)}+b_j)$, \quad $j=1,\ldots,J$
\STATE \textbf{return} $\hat{\mathbf{y}}=[\hat{y}_1,\ldots,\hat{y}_J]$

\end{algorithmic}
\end{algorithm}

%% file: section/add_phq_dataset.tex
\section{Construction and evaluation of \longcounsel}\label{apdx:longcounsel}

\longcounsel is constructed as a controlled multi-session counseling corpus for current-session PHQ-8 prediction with optional prior-session context. The construction follows a four-stage process that (1) defines a stable client profile and session plan, (2) assigns a five-visit depression trajectory, (3) decomposes each visit-level severity anchor into eight PHQ-8 symptom scores, and (4) renders those private symptom states as natural counseling dialogue. Empirical sources and deterministic validation rules guide each step.

\subsection{Construction}
\label{apdx:longcounsel-construction}

\paragraph{Generation chain.}
Each trajectory contains five counseling sessions for one client. For session $t$, the latent symptom state is a symptom-level PHQ-8 vector
\begin{equation}
    \mathbf{y}^{(t)} = [y_1^{(t)}, \ldots, y_8^{(t)}], \qquad
    y_j^{(t)} \in [0, 3],
\end{equation}
whose sum gives the visit-level PHQ-8 total score. PsychEval-derived cases~\citep{pan2026psycheval} provide stable client backgrounds, presenting concerns, and session-stage structures. PSYCHE-D~\citep{Makhmutova2021PredictingCI} provides the longitudinal severity anchors, aligning the five-visit changes with empirical depression trajectories. NHANES DPQ responses~\citep{cdc_nhanes_dpq_l_2024} then provide observed symptom co-occurrence patterns. For each anchor, we sample a complete PHQ-style response vector and retain the first eight symptoms as the latent PHQ-8 state.

The client-profile step deterministically converts PsychEval-derived cases. It preserves stable identity, biography, presenting concern, topic, growth history, beliefs, and target behaviors, while removing counselor-only administrative material and process instructions. Session themes are rewritten as client-side disclosure beats that specify a concrete event, bodily or contextual triggers, feared meaning, avoidance cost, and desired help. Profiles without valid client information or session plans are removed before trajectory assignment.

For trajectory reconstruction, PSYCHE-D quarterly observations are chained into five anchors. The first visit uses the first available start score, and subsequent visits use the corresponding quarter-end scores. When adjacent quarterly windows disagree on a shared boundary, the two boundary values are averaged and rounded. For each reconstructed PHQ-9 anchor, NHANES DPQ rows with complete symptom responses are grouped by total score. A deterministic hash of the global seed, participant id, visit index, and anchor score selects one matching response vector. The first eight PHQ symptoms form the PHQ-8 vector, and their sum becomes the latent PHQ-8 total. The decomposition uses exact observed NHANES total-score buckets, with all-zero and all-maximum anchors handled as deterministic boundary cases.

\paragraph{Condensed prompt templates.}
The following templates are used only after deterministic profile and trajectory construction. The labels identify the generation stages and appear outside the prompt text. Bracketed spans denote instance-specific content instantiated from one client's profile, one session's private symptom cues, the public dialogue history, retrieved memories, and the current turn plan.

\noindent\textbf{Behavior-cue generation.}
\begin{quote}\small
Given a PHQ-8 symptom, an ordinal severity level, and the intended direction when applicable, write short candidate behavioral cues that could naturally appear in a counseling client. Each cue must describe one lived-experience signal in 12--30 words. Do not mention PHQ, questionnaires, scores, symptom names, response options, or measurement language. Do not write first-person questionnaire answers or direct paraphrases of the PHQ symptom. Avoid cues that primarily express another PHQ-8 symptom.
\end{quote}

\noindent\textbf{Counselor-turn generation.}
\begin{quote}\small
You are the counselor. Given the public conversation so far, the session focus, a brief carryover note, relevant past-session memories, the target counseling move for this turn, and the RealCBT-derived turn form, write only the next counselor utterance in English. Follow the specified surface form, continue the session naturally, avoid repetition and premature closure, and do not mention latent states, PHQ, scores, or hidden generation instructions. Return only the visible counselor message in the required structured form.
\end{quote}

\noindent\textbf{Client-turn generation.}
\begin{quote}\small
You are the client described by the private profile. Given the public conversation so far, the current session disclosure beat, private symptom cues, stable background, carryover note, relevant memories, and the RealCBT-derived turn form, write only the next client utterance in English. Express the private cues indirectly through ordinary speech, partial disclosure, hesitation, affect, or concrete examples. Do not name PHQ, scores, questionnaire symptoms, cue lists, or hidden instructions; do not speak like a counselor. Return the visible client message and a short emotion phrase in the required structured form.
\end{quote}

\noindent\textbf{Self-report generation.}
\begin{quote}\small
After the session, answer PHQ-8 as the same client using the private symptom cues, stable profile, carryover note, and completed transcript. For each shuffled symptom, choose exactly one of the provided shuffled option texts. Base the answer on what the client would endorse, not on what was explicitly disclosed. Return only the required structured response, with no rationale, commentary, paraphrased options, or extra text.
\end{quote}

\paragraph{From symptom scores to dialogue.}
Each PHQ-8 symptom score is translated into private behavioral cues for dialogue generation. The behavior-cue prompt generates short, atomic, naturalistic cues for each PHQ-8 symptom--score cell. It emphasizes lived experience over questionnaire wording and excludes PHQ names, option labels, numeric scores, direct symptom paraphrases, first-person self-report phrasing, and strong cross-symptom contamination. These cues cover affect, functioning, cognition, somatic burden, and interaction style. Candidate cues are first checked within their target cell and then pruned under joint composition tests that sample full PHQ-8 profiles and retain cues whose induced self-reports recover the intended symptom and total scores. After pruning, each PHQ-8 symptom--score cell contains ten validated behavioral cues. At generation time, each symptom contributes five cues. Fractional scores are represented by sampling from adjacent integer severity levels in proportion to the fractional value. The client simulator receives these cues as private self-beliefs and is explicitly instructed not to reveal questionnaire language or numeric labels.

The conversation generator conditions on the private symptom cues, the PsychEval-derived client profile, memory from earlier visits, and turn-level constraints calibrated from RealCBT~\citep{Wang2025FeelTD}. Counselor prompts receive only public dialogue context, a short carryover note, retrieved memories, the session focus, the current counseling-move target, and a RealCBT-derived turn-form card. Client prompts receive the public context plus private symptom cues, stable background, session beat guidance, carryover, retrieved memories, and the same style of turn-form card. RealCBT provides surface calibration. Speaker- and stage-specific statistics determine sentence-count patterns, turn length, question form, discourse starters, and lexical overlap, while the client profile and symptom cues determine the clinical content. These constraints embed symptom cues in counseling exchanges.

In the main set, visible counselor and client turns are generated with Qwen3.5-35B-A3B under moderate sampling, using temperature $0.6$, nucleus mass $0.95$, top-$k$ sampling at $20$, and a 512-token cap per turn. Longitudinal run assignment uses seed $42$, and the inference backend uses a fixed server seed. Each run contains five sessions with 20 counselor turns and 20 client turns per session. Generated turns must be in English, role-consistent, non-empty, non-repetitive, and free of questionnaire or score leakage and premature closure. Violations trigger targeted repair attempts.

\paragraph{Latent target and self-report supervision.}
In \longcounsel, the latent PHQ-8 state is known by construction and serves as the controlled evaluation target. Models train on simulated self-report scores and are evaluated against the latent state that drives dialogue generation. The latent values serve as private generation conditions, while the visible dialogue conveys symptom evidence through client behavior and language.

After each completed session, the client simulator privately completes PHQ-8 using the same behavior cues, stable profile, carryover note, and transcript. Symptom and option order are deterministically shuffled for each pass, and the prompt requires exact option texts without rationale or extra prose. We generate five deterministic self-report passes and average them symptomwise. Invalid structure, missing symptoms, or paraphrased options trigger a repair prompt. Sessions without a valid self-report are excluded from split construction. Session summaries and self-report responses use deterministic decoding, with 1024 tokens allowed for summaries and 1024--1536 tokens for self-report responses depending on the generation setting.

\paragraph{Dataset statistics and use.}
The complete main Qwen-generated set used in all reported experiments contains 3,690 five-session trajectories and 18,450 labeled counseling sessions. A trajectory-level audit assigns generated clients to fluctuating, stable, improving, and worsening patterns, with all four classes represented. All model experiments use trajectory-level partitions, keeping sessions from each trajectory in one partition.

Filtering precedes split construction. Trajectory reconstruction keeps strict five-visit PSYCHE-D participants, NHANES decomposition uses only complete valid symptom-response rows, the cue bank is pruned to a uniform ten cues per symptom--score cell, and generated runs must contain a valid transcript, latent target, and simulated self-report. The main-set split manifest is generated over the resulting 3,690 usable trajectories, assigning 2,583 to training, 369 to validation, and 738 to testing. Split assignment is produced by sorting run ids, applying a seed-42 random shuffle within the \longcounsel parent group, and assigning the 0.7/0.1/0.2 train/validation/test ratios at the trajectory level. The manifest records the split for each run id.

\longcounsel is used to evaluate current-session PHQ-8 prediction under prescribed symptom trajectories, partial disclosure, and cross-session continuity. We therefore pair experiments on \longcounsel with evaluation on \daic~\citep{Gratch2014TheDA}, which provides a real single-session anchor for text-based depression assessment.

\subsection{Self-report fidelity audit}
\label{apdx:longcounsel-evaluation}

\begin{figure}[!htbp]
    \centering
    \subfloat[Item, profile, and total-score exact-match rates.\label{fig:cue-composition-accuracy}]{%
  \includegraphics[width=0.31\linewidth]{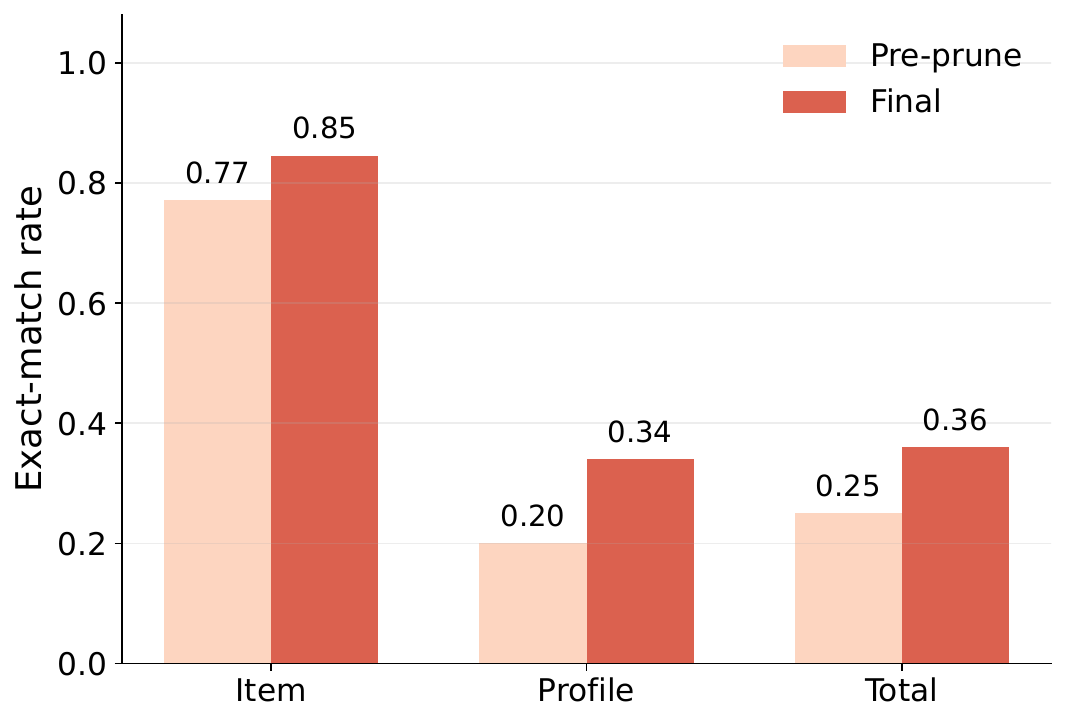}%
}
    \hfill
    \subfloat[Item and total-score MAE.\label{fig:cue-composition-error}]{%
  \includegraphics[width=0.31\linewidth]{fig/cue_composition_error.pdf}%
}
    \hfill
    \subfloat[MAE for each PHQ-8 symptom.\label{fig:cue-symptom-mae}]{%
  \includegraphics[width=0.31\linewidth]{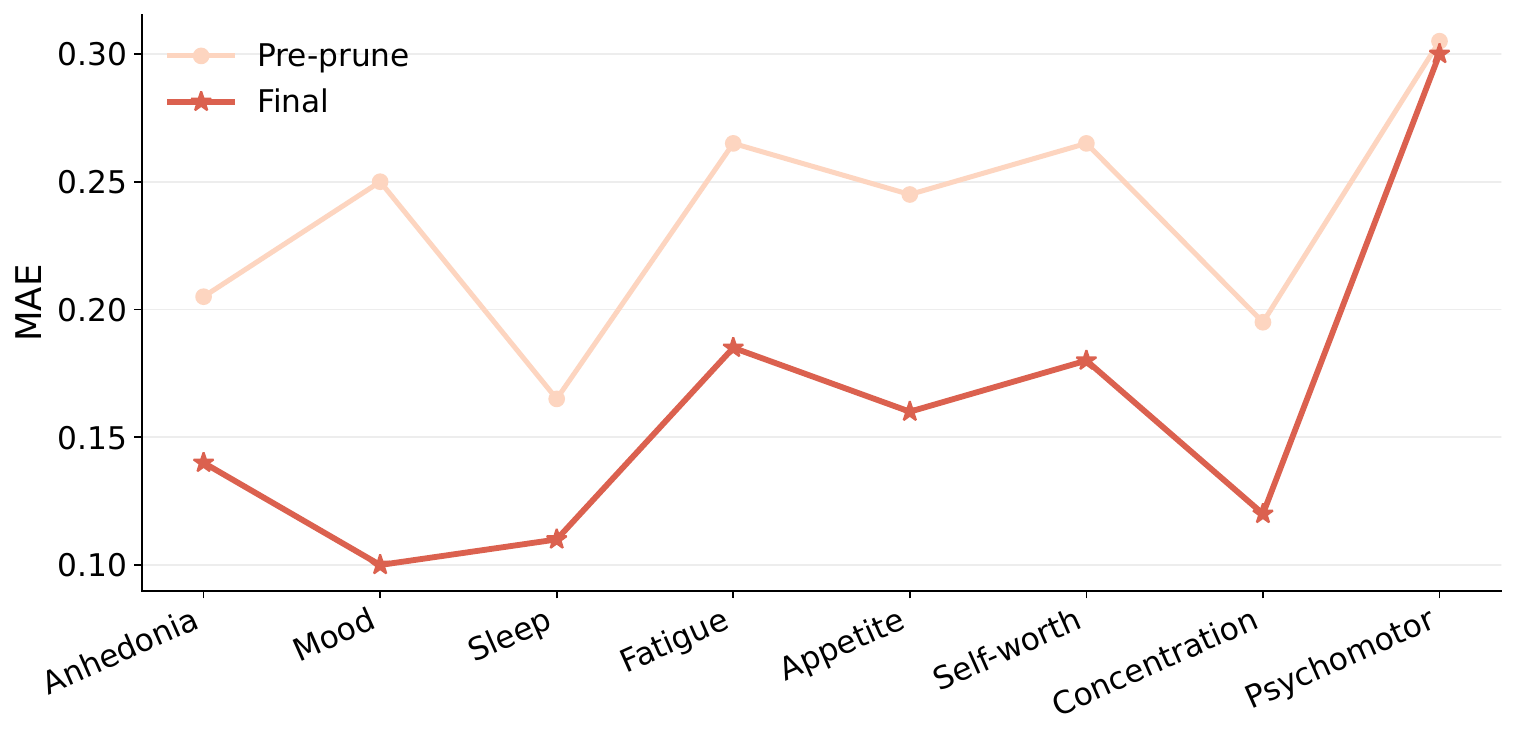}%
}
    \caption{Self-report fidelity before and after cue pruning, with five cues per symptom.}
    \label{fig:cue-composition}
\end{figure}

Pruning the candidate behavior-cue bank substantially improves agreement between simulated self-reports and prescribed latent PHQ-8 states across all granularity levels. Figure~\ref{fig:cue-composition} compares pre-pruning and final cue banks under the runtime composition protocol (five sampled cues per symptom), reporting exact-match rates, aggregate item and total-score MAE, and per-symptom MAE against the hidden PHQ-8 vector. Pruning raises exact-match rates by $8$ percentage points at the item level, $14$ points at the full-profile level, and $11$ points at the total-score level, while reducing item MAE by $31.6\%$, total-score MAE by $28.4\%$, and error across every individual PHQ-8 symptom. Section~\ref{subsec:dataset-validation} and Figure~\ref{fig:main-cue-error} summarize this fidelity audit.

\subsection{RealCBT similarity audit}
\label{apdx:longcounsel-realcbt}
\begin{figure}[htbp]
    \centering
    \subfloat[Aggregate psycholinguistic quantile distance.\label{fig:psycholinguistic-overall}]{%
  \includegraphics[width=0.31\linewidth]{fig/psycholinguistic_overall_distance.pdf}%
}
    \hfill
    \subfloat[Judge distance from chance.\label{fig:realcbt-indistinguishability}]{%
  \includegraphics[width=0.31\linewidth]{fig/realcbt_indistinguishability.pdf}%
}
    \hfill
    \subfloat[Feature-level quantile distance.\label{fig:feature-distance}]{%
  \includegraphics[width=0.31\linewidth]{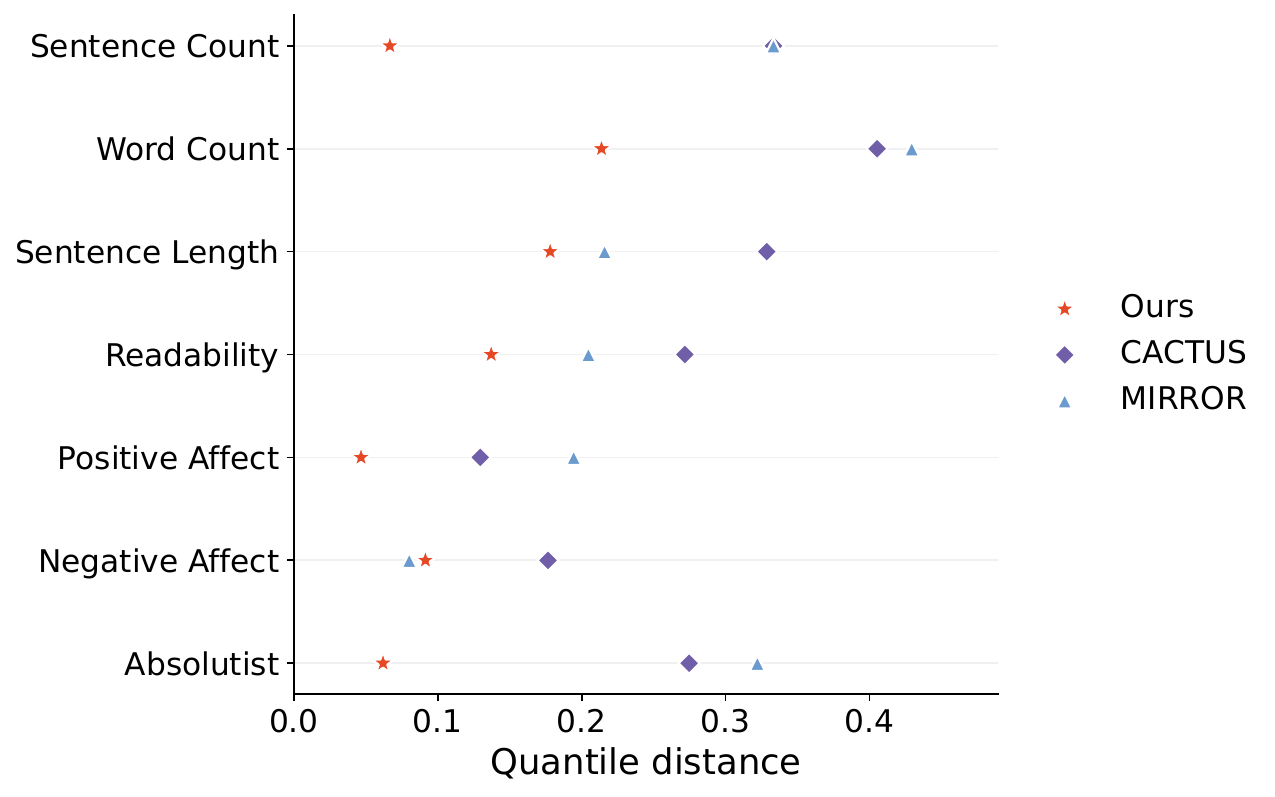}%
}
    \caption{Dialogue similarity to RealCBT for \longcounsel, CACTUS, and MIRROR. Lower values indicate closer alignment under each measure.}
    \label{fig:realcbt-audit}
\end{figure}

\longcounsel matches the surface and affective characteristics of real counseling dialogue in RealCBT more closely than existing synthetic counseling corpora~\citep{Wang2025FeelTD,lee2024cactus,kim2025mirror}. Figure~\ref{fig:realcbt-audit} evaluates alignment with RealCBT for \longcounsel, CACTUS, and MIRROR using aggregate psycholinguistic quantile distance, pointwise judge distance from chance (accounting for systematic judgment inversion), and feature-level quantile distances over speaker-specific turn length, sentence structure, readability, affective language, and absolutist language. \longcounsel reduces aggregate psycholinguistic distance by $58.4\%$ relative to CACTUS and $55.1\%$ relative to MIRROR, lowers judge distance from chance by $13.2\%$ and $27.9\%$, respectively, and achieves the closest RealCBT alignment on six of the seven psycholinguistic features (with MIRROR slightly closer only on negative-affect ratio). Section~\ref{subsec:dataset-validation} and Figures~\ref{fig:main-psycholinguistic}--\ref{fig:main-judge-distance} highlight these comparisons in the main paper.

%% file: section/add_experiment_setting.tex
\section{Additional experiment settings}
\label{sec:add_experiment_setting}

\paragraph{Hyperparameters.}
Table~\ref{tab:hyperparams} reports the reference configuration selected on the main \longcounsel validation split. The model in this configuration has approximately $0.972\text{M}$ trainable parameters. On LongCounsel-8-GPT54-Mini, memory uses $S{=}16$ slots, history dropout is $p_{\mathrm{hist}}{=}0.2$, and $\lambda_{\mathrm{sym}}{=}0$. On \daic, the encoder has three layers, $\lambda_{\mathrm{sym}}{=}0$, and memory is bypassed because no prior session is available. The retained-client-turn cap is $N_{\max}{=}80$ on \daic. Every longitudinal session contains 20 client turns and fits within this cap. The frozen \textsc{E-DAIC} transfer evaluation retains the \daic-trained model and its selected settings. All default comparisons disable the false-negative penalty ($\lambda_{\mathrm{FN}}{=}0$). Appendix~\ref{subsec:fn-sweep} reports the separate optional-penalty experiment.
\begin{table}[ht]
\centering
\caption{EmoTrack reference hyperparameters for the main \longcounsel set, with the \daic client-turn cap also listed.}
\label{tab:hyperparams}
\small
\begin{tabular}{@{}llll@{}}
\toprule
\textbf{Component} & \textbf{Value} & \textbf{Component} & \textbf{Value} \\
\midrule
Model dimension $d$ & 64 
& AIDA features $F$ & 23 \\
Attention heads $h$ & 4 
& Dropout & 0.1 \\
Feed-forward dimension & 256 
& History dropout $p_{\mathrm{hist}}$ & 0.1\\
Encoder layers $L_{\mathrm{enc}}$ & 2 
& Learning rate & $10^{-3}$ \\
Decoder layers $L_{\mathrm{dec}}$ & 4 
& Weight decay & $10^{-2}$ \\
Memory slots $S$ & 4
& Batch size & 32 \\
Score MLP hidden dim & 32 
& Warmup ratio & 0.05 \\
DAIC max client turns $N_{\max}$ & 80
& Symptom-loss weight $\lambda_{\mathrm{sym}}$ & 0.5 \\
PHQ-8 items $J$ & 8
& False-negative weight $\lambda_{\mathrm{FN}}$ & 0 \\
\bottomrule
\end{tabular}
\end{table}

\paragraph{Training.}
We train with AdamW~\citep{loshchilov2019decoupled} (lr${}=10^{-3}$, weight decay${}=10^{-2}$), batch size~32, linear warmup over the first 5\% of steps followed by cosine decay, gradient clipping at norm~1.0, and early stopping with patience~8 on validation loss. Configurations are selected by validation MAE as specified in Section~\ref{subsec:experiment-settings}. The memory configuration uses the immediately preceding session, with longer scopes evaluated separately in Appendix~\ref{subsec:history-scope}.
Training takes approximately 5 minutes per seed on a single GPU.

\paragraph{Environment.}
All experiments were performed on a server running Ubuntu 22.04 LTS, equipped with an AMD EPYC 9654 96-core processor (192 threads), 1.1\,TiB of system memory, and 4 NVIDIA H100 PCIe GPUs with 80\,GiB HBM each (320\,GiB aggregate GPU memory).

\paragraph{Baseline implementations.}
Our baseline reproductions match or improve upon published clinical-based results on \daic, while training-based baselines exhibit higher error and seed sensitivity under the official split. Table~\ref{tab:baseline-comparison-daic} reports the five-seed test MAE on \daic for all reproduced baselines using released algorithms or author code. For clinical-based baselines, our five-seed mean is $3.0\%$ below the $2.91$ MAE reported for AIDA~\citep{lee2026interpretable} and $12.5\%$ below the $3.46$ MAE reported for LMIQ~\citep{rosenman2024llm}.

For training-based baselines, checkpoint and initialization sensitivity on the small \daic split leads to higher multi-seed error than single-run published scores. Relative to the $3.80$ test MAE reported by Lau et al.~\citep{lau2023automatic}, our released dual-encoder run with all-mpnet-base-v2 has a five-seed mean $16.8\%$ higher—spanning from $4.4\%$ above on its best seed to $42.6\%$ above on a seed that selects a near-constant epoch-0 checkpoint—while the Qwen3-Embedding variant averages $7.7\%$ above the published score with a best seed $4.4\%$ below it. Relative to the $3.78{\pm}0.13$ test MAE reported by Milintsevich et al.~\citep{milintsevich2023towards}, our five-seed mean is $17.4\%$ higher (with the best seed $13.6\%$ higher and a direct author-repository audit $19.5\%$ higher), reflecting an author pipeline that excludes one training participant with missing item labels and two corrupted development transcripts before selecting checkpoints by development loss.

Adapting EnsemBERT~\citep{ravenda2025transforming} from social-media BDI-II prediction to clinical PHQ-8 estimation requires hyperparameter tuning to prevent constant-output collapse. We evaluate test MAE on \daic before and after a 729-configuration grid search (six optimization and capacity hyperparameters over three values each across five seeds) on the official training and validation splits, keeping the architecture, frozen representation model, retrieval model, and loss weights fixed. Whereas two of the five all-mpnet-base-v2 seeds collapse to all-zero predictions under the published settings, the selected configuration (learning rate $10^{-3}$, weight decay $10^{-4}$, 30 retrieved posts, 256 dense units, two attention heads, and batch size 64) eliminates constant-output seeds and reduces five-seed test MAE by $27.4\%$ in Table~\ref{tab:baseline-comparison-daic}, and is applied unchanged across all datasets and embedding variants.

%% file: section/add_experiment.tex
\section{Additional experiments}
\label{apdx:add_experiment}

This appendix reports the complete cross-dataset method comparison, extended component and hyperparameter ablations, clinical screening and per-symptom bias analyses, longitudinal transition and trajectory-type stratifications, and inference runtime benchmarks. Unless otherwise noted, results are the five-seed mean $\pm$ sample standard deviation of test MAE across seeds $0,\ldots,4$, where \textsuperscript{\ensuremath{\dagger}} marks the configuration selected by validation MAE and bold denotes the best test mean within each column.

\subsection{Complete method comparison}
\label{subsec:all-results}

EmoTrack delivers the most consistent accuracy across single-session, transfer, and longitudinal regimes, bridging the gap between clinical-based and training-based baselines. Table~\ref{tab:all-results} reports the five-seed total-score test MAE across all five evaluation sets. Whereas clinical-based methods lead the baselines on real interviews and training-based methods lead on synthetic corpora, EmoTrack ranks first on \daic ($12.5\%$ lower MAE than AIDA) and the complementary set ($1.9\%$ lower than EnsemBERT with Qwen3-Emb), ranks second under \textsc{E-DAIC} transfer ($6.5\%$ behind LMIQ and at least $24.6\%$ below all trained baselines), and remains within $1.1\%$ and $5.2\%$ of the top trained predictors on \longcounsel and LongCounsel-8-Luna. Section~\ref{subsec:performance-comparison} and Figure~\ref{fig:cross-regime-radar} summarize the aggregate comparison.

\input{table/all_results}

\subsection{Ablation study: encoder--decoder layers}
\label{subsec:ablation-encoder-decoder-layers}

Prediction accuracy exhibits modest sensitivity to encoder--decoder depth across datasets. Table~\ref{tab:ablation-encoder-decoder-layers} reports total-score test MAE across five transformer depth configurations on \longcounsel, \daic, and the complementary set. Test means vary by at most $1.9\%$ across depths on \longcounsel (with the validation-selected 2--4 configuration within $0.6\%$ of the minimum), while the selected 3--4 configuration achieves the lowest error on \daic (up to $6.0\%$ below other depths) and the selected 2--4 configuration remains within $1.2\%$ of the test minimum on the complementary set. Section~\ref{subsec:backbone} describes the backbone architecture.

\input{table/ablations/table_encoder_decoder_layers}

\subsection{Ablation study: history dropout}
\label{subsec:ablation-history-dropout}

Light history dropout improves longitudinal generalization on the full-scale benchmark, whereas aggressive dropout degrades accuracy on both datasets. Table~\ref{tab:ablation-history-dropout} evaluates total-score test MAE across history-dropout rates $p_{\mathrm{hist}}\in\{0.0,0.1,0.2,0.3\}$ on \longcounsel and the complementary set. On \longcounsel, the validation-selected rate of $0.1$ achieves the lowest error and reduces MAE by $2.2\%$ relative to no dropout; on the complementary set, no dropout yields the lowest test error ($1.4\%$ below the validation-selected $0.2$), while increasing dropout to $0.3$ raises MAE by $1.9\%$ and $4.1\%$ relative to the best setting on each dataset. Figure~\ref{fig:history-context-ablation} visualizes these trends in the main paper.

\input{table/ablations/table_history_dropout}

\subsection{Ablation study: retained client turns}
\label{subsec:ablation-max-context-sentences}

Bounding the number of retained client turns improves prediction on long clinical interviews by filtering late-session noise without truncating core disclosures. Table~\ref{tab:ablation-max-context-sentences} reports total-score test MAE on \daic as the client-turn cap $N_{\max}$ varies from 20 to 384. The validation-selected cap of 80 client turns achieves the lowest error, reducing MAE by $4.6\%$ relative to 384 client turns and by $9.3\%$--$10.4\%$ relative to the shorter 40- and 20-turn caps, while still retaining all 20 client turns in every longitudinal session. Figure~\ref{fig:history-context-ablation} plots this sweep alongside history dropout.

\input{table/ablations/table_max_context_sentences}

\subsection{Ablation study: memory mechanism}
\label{subsec:ablation-memory-mechanism}

Combining summary and slot-retrieval memory provides the most reliable gains across dialogue generators, even though summary-only memory suffices on the main corpus. Table~\ref{tab:ablation-memory-mechanism} compares total-score test MAE on \longcounsel and the complementary set across memory pathway configurations. Relative to no previous memory, the validation-selected combined pathway reduces MAE by $3.2\%$ on \longcounsel and $0.5\%$ on the complementary set, whereas summary-only memory lowers MAE by $3.9\%$ on \longcounsel ($0.8\%$ below the combined pathway) but increases error by $1.9\%$ on the complementary set. Section~\ref{subsec:memory} details both pathways.

\input{table/ablations/table_memory_mechanism}

\subsection{Ablation study: memory slots}
\label{subsec:ablation-memory-slots}

Compact slot capacity achieves the lowest test error on both longitudinal benchmarks, preventing over-parameterization of prior-session context. Table~\ref{tab:ablation-memory-slots} reports total-score test MAE as the number of memory slots $S$ varies from 4 to 16 on \longcounsel and the complementary set. Four slots yield the lowest test error on both datasets, whereas expanding capacity to 16 slots increases MAE by $2.8\%$ on \longcounsel and by $1.0\%$ (the validation-selected setting, within a $1.9\%$ overall range) on the complementary set. Section~\ref{subsec:memory} defines the slot-compression module.

\input{table/ablations/table_memory_slots}

\subsection{Ablation study: history scope}
\label{subsec:history-scope}

Extending memory to two prior sessions achieves the lowest error across both longitudinal datasets, outperforming both single-session and unbounded history. Table~\ref{tab:history-scope} compares total-score test MAE across historical scopes when feeding the selected prior client-turn embeddings into the unchanged slot compressor. Relative to one prior session, two-session history reduces MAE by $1.4\%$ on \longcounsel and $2.3\%$ on the complementary set ($4.5\%$ and $2.8\%$ below no memory), whereas compressing all prior sessions yields smaller reductions of $0.5\%$ and $1.5\%$. Section~\ref{subsec:memory} describes the default single-session memory configuration.

\input{table/analysis/history_scope}

\subsection{Ablation study: symptom-wise loss}
\label{subsec:ablation-item-loss}

Auxiliary symptom-level supervision substantially improves item-level accuracy and slightly lowers total-score error when training data is abundant, whereas aggregate-only supervision is preferred on smaller datasets. Table~\ref{tab:ablation-item-loss-total-item-mae} reports total-score and symptom-level test MAE across symptom-loss weights $\lambda_{\mathrm{sym}}\in\{0,0.1,0.3,0.5,1.0\}$. On \longcounsel, the validation-selected $\lambda_{\mathrm{sym}}=0.5$ reduces symptom MAE by $25.1\%$ and total-score MAE by $0.9\%$ relative to aggregate-only training ($\lambda_{\mathrm{sym}}=0$), with $\lambda_{\mathrm{sym}}=1.0$ reducing symptom MAE by a further $1.7\%$; conversely, $\lambda_{\mathrm{sym}}=0$ achieves the lowest error on \daic ($3.5\%$--$8.3\%$ below nonzero weights) and is also selected by validation on the complementary set ($2.6\%$ above the test minimum at $\lambda_{\mathrm{sym}}=0.3$). Section~\ref{subsec:training-objective} defines the composite loss.

\input{table/ablations/table_item_loss_total_item_mae}

\subsection{Clinical-feature extractor}
\label{subsec:extractor-swap}

EmoTrack consistently outperforms clinical-based regressors across different upstream LLM feature extractors. We evaluate five-seed test MAE on \daic under Qwen3.5-35B-A3B and GPT-5.6 Luna clinical extraction using a shared reference configuration (two encoder layers, four decoder layers, $\lambda_{\mathrm{sym}}{=}0.5$, and a cap of 80 client turns). Under GPT-5.6 Luna extraction, EmoTrack ($2.6548{\pm}0.1057$) reduces MAE by $5.5\%$ relative to matched AIDA ($2.8091{\pm}0.1003$) and by $15.0\%$ relative to LMIQ ($3.1223{\pm}0.0508$); under Qwen extraction, EmoTrack ($2.5831{\pm}0.2332$) reduces MAE by $8.5\%$ relative to AIDA ($2.8234{\pm}0.2099$) and by $14.7\%$ relative to LMIQ ($3.0285{\pm}0.0460$), with EmoTrack's own mean error varying by only $2.8\%$ between extractors. Section~\ref{subsec:input-representation} describes the clinical-feature token representation.

\subsection{False-negative penalty}
\label{subsec:fn-sweep}

Enabling the optional false-negative penalty in Eq.~\eqref{eq:fn-aware-loss} substantially lowers missed positive screens on the complementary dataset at a moderate cost in false positives and total-score error. Table~\ref{tab:fn-aware-comparison} reports test FNR, FPR, and total-score MAE at the clinical screening cutoff of $\text{PHQ-8}\geq10$ across all baselines and EmoTrack with $\lambda_{\mathrm{FN}}\in\{0,2.5\}$. Selected by validation FNR over $\lambda_{\mathrm{FN}}\in\{0,0.5,1,2,2.5,3,4,5\}$, $\lambda_{\mathrm{FN}}=2.5$ achieves the lowest FNR among all evaluated methods—reducing FNR by $40.4\%$ relative to $\lambda_{\mathrm{FN}}=0$ and by $28.2\%$ relative to the strongest baseline, Lau et al.—while increasing FPR by $15.6$ percentage points and MAE by $5.0\%$. Section~\ref{subsec:optional-fn} and Figure~\ref{fig:optional-fn} summarize this screening trade-off.

\input{table/analysis/fn_aware_comparison}

Across the penalty-weight sweep, sensitivity gains saturate at $\lambda_{\mathrm{FN}}=2.5$ while larger weights degrade score accuracy without further reducing missed positives. Relative to $\lambda_{\mathrm{FN}}=0$, FNR already drops by $19.8\%$ at $\lambda_{\mathrm{FN}}=0.5$ before reaching its $40.4\%$ maximum reduction at $\lambda_{\mathrm{FN}}=2.5$; for $\lambda_{\mathrm{FN}}\in\{3,4,5\}$, FNR rebounds by $3.3\%$--$14.7\%$ above the $\lambda_{\mathrm{FN}}=2.5$ minimum while total-score MAE rises by up to $14.1\%$ above the unpenalized model.

\subsection{Per-item error and directional bias}
\label{subsec:symptom-bias}

EmoTrack maintains sub-point accuracy across all eight PHQ-8 symptoms on real clinical interviews, with opposing item-level biases that partially cancel at the total-score level. Table~\ref{tab:symptom-bias} reports per-symptom MAE and mean signed bias (prediction minus reference label) on the 35-participant \daic development split with released item annotations. Every symptom remains at least $24.6\%$ below a one-point MAE—with core affective symptoms (anhedonia and depressed mood) averaging $27.8\%$ lower error than somatic and cognitive symptoms—while negative biases on sleep, fatigue, and appetite offset positive biases on psychomotor change and anhedonia, whose two largest magnitudes match within $0.7\%$. Section~\ref{subsec:training-objective} describes the item-level prediction head.

\input{table/analysis/symptom_bias}

\subsection{Current-score error after session-to-session transitions}
\label{subsec:transitions}

Historical context improves estimation most when symptom severity persists across visits. We stratify test sessions by the absolute change in latent PHQ-8 total score from the preceding session and compare total-score MAE for EmoTrack with and without one-session memory. When severity is unchanged, memory reduces MAE by $9.2\%$ on \longcounsel and $5.9\%$ on the complementary set, whereas after shifts of at least five points, memory increases MAE by $2.2\%$ and $3.4\%$, respectively. Section~\ref{subsec:memory} and Appendix~\ref{subsec:trajectory-strat} connect this transition behavior to memory design and trajectory-level performance.

Two real-data longitudinal studies support the premise that temporal history carries predictive signal. In a 12-week cohort of 229 patients with major depressive disorder, adding dynamic longitudinal features raised the best overall trajectory-classification accuracy from below $50\%$ to $60.87\%$~\citep{Zhong2026ContributionOL}. In a separate year-long study, initial and intermediate PHQ-9 states were the most important predictors of later deterioration~\citep{Makhmutova2021PredictingCI}.

\subsection{Current-score error by trajectory type}
\label{subsec:trajectory-strat}

EmoTrack achieves the lowest error on stable depression trajectories and remains closely competitive on improving trajectories, while rapid symptom shifts favor memory-free encoders. Table~\ref{tab:trajectory-strat} stratifies five-seed test MAE on \longcounsel across improving, worsening, stable, and fluctuating PHQ-8 trajectory types. On stable trajectories—the largest clinical subgroup—EmoTrack ranks first and reduces MAE by $5.8\%$ relative to the strongest baseline, Lau et al.\ (Qwen3-Emb), while trailing that baseline by only $1.3\%$ on improving trajectories ($7.0\%$ on fluctuating and $10.3\%$ on worsening trajectories) and outperforming all clinical-based methods across every group. Appendix~\ref{subsec:transitions} examines the corresponding session-to-session transition effects.

\input{table/analysis/trajectory_strat}

\subsection{Runtime and inference cost}
\label{subsec:runtime-compute-cost}

With cached clinical features and client-turn embeddings, EmoTrack completes downstream test-set inference in under two minutes per benchmark, avoiding the multi-hour cost of repeated transcript encoding. Table~\ref{tab:baseline-timing} compares single-GPU inference wall-clock hours on \longcounsel and \daic across all evaluated methods (excluding upstream feature extraction and training). Whereas hierarchical and dual-encoder baselines~\citep{milintsevich2023towards,lau2023automatic} repeatedly encode full transcripts, EmoTrack reduces inference time by $99.5\%$--$99.9\%$ on \longcounsel and $98.4\%$--$99.4\%$ on \daic while remaining within seconds of lightweight tabular regressors. Section~\ref{subsec:experiment-settings} and Appendix~\ref{sec:add_experiment_setting} detail the hardware and evaluation protocol.

\input{table/methods_timing}

%% file: table/all_results.tex
\begin{table}[ht]
\centering
\small
\caption{Total-score MAE across all evaluation sets, with the lowest mean in each column in bold. Values are mean $\pm$ sample standard deviation over five seeds.}
\label{tab:all-results}
\begin{tabular}{@{}lccccc@{}}
\toprule
\multirow{2}{*}{\textbf{Method}} & \multicolumn{5}{c}{\textbf{Overall MAE $\downarrow$}} \\
\cmidrule(l){2-6}
& \longcounsel & \textsc{LC8-Luna} & \makecell{\textsc{GPT-5.4-mini}\\\textsc{Comp.}} & \daic & \makecell{\daic\\$\rightarrow$\,\textsc{E-DAIC}} \\
\midrule
\makecell[l]{AIDA\\(now)} & \makecell{3.2210\\{\small $\pm$ 0.0068}} & \makecell{3.4932\\{\small $\pm$ 0.0045}} & \makecell{4.0689\\{\small $\pm$ 0.0133}} & \makecell{2.8234\\{\small $\pm$ 0.2099}} & \makecell{4.2538\\{\small $\pm$ 0.2697}} \\
\makecell[l]{AIDA\\(all)} & \makecell{3.2517\\{\small $\pm$ 0.0051}} & \makecell{3.5173\\{\small $\pm$ 0.0083}} & \makecell{4.0948\\{\small $\pm$ 0.0243}} & -- & -- \\
\makecell[l]{LMIQ\\(now)} & \makecell{2.9792\\{\small $\pm$ 0.0117}} & \makecell{3.2784\\{\small $\pm$ 0.0205}} & \makecell{4.0095\\{\small $\pm$ 0.0566}} & \makecell{3.0285\\{\small $\pm$ 0.0460}} & \makecell{\textbf{3.4722}\\{\small \textbf{$\pm$ 0.0729}}} \\
\makecell[l]{LMIQ\\(all)} & \makecell{2.9585\\{\small $\pm$ 0.0185}} & \makecell{3.2285\\{\small $\pm$ 0.0146}} & \makecell{3.9908\\{\small $\pm$ 0.0513}} & -- & -- \\
\makecell[l]{Milintsevich\\et al.} & \makecell{2.6953\\{\small $\pm$ 0.0211}} & \makecell{\textbf{2.5588}\\{\small \textbf{$\pm$ 0.0226}}} & \makecell{4.1530\\{\small $\pm$ 0.0602}} & \makecell{4.4370\\{\small $\pm$ 0.1197}} & \makecell{5.0497\\{\small $\pm$ 0.1233}} \\
EnsemBERT & \makecell{2.9092\\{\small $\pm$ 0.1128}} & \makecell{3.0336\\{\small $\pm$ 0.0394}} & \makecell{4.0038\\{\small $\pm$ 0.1396}} & \makecell{4.1149\\{\small $\pm$ 0.3339}} & \makecell{4.9047\\{\small $\pm$ 0.3487}} \\
\makecell[l]{EnsemBERT\\(Qwen3-Emb)} & \makecell{2.7492\\{\small $\pm$ 0.0343}} & \makecell{3.0561\\{\small $\pm$ 0.3747}} & \makecell{3.9184\\{\small $\pm$ 0.0753}} & \makecell{4.8255\\{\small $\pm$ 0.7277}} & \makecell{5.2465\\{\small $\pm$ 0.5317}} \\
Lau et al. & \makecell{2.6204\\{\small $\pm$ 0.0194}} & \makecell{2.5768\\{\small $\pm$ 0.0464}} & \makecell{4.0689\\{\small $\pm$ 0.0865}} & \makecell{4.4367\\{\small $\pm$ 0.5744}} & \makecell{5.1880\\{\small $\pm$ 0.3158}} \\
\makecell[l]{Lau et al.\\(Qwen3-Emb)} & \makecell{\textbf{2.5944}\\{\small \textbf{$\pm$ 0.0232}}} & \makecell{2.5985\\{\small $\pm$ 0.0499}} & \makecell{4.0640\\{\small $\pm$ 0.1034}} & \makecell{4.0927\\{\small $\pm$ 0.2795}} & \makecell{5.3641\\{\small $\pm$ 0.2134}} \\
\midrule
\textbf{EmoTrack} & \makecell{2.6238\\{\small $\pm$ 0.0610}} & \makecell{2.6913\\{\small $\pm$ 0.0362}} & \makecell{\textbf{3.8447}\\{\small \textbf{$\pm$ 0.1181}}} & \makecell{\textbf{2.4708}\\{\small \textbf{$\pm$ 0.3892}}} & \makecell{3.6977\\{\small $\pm$ 0.2665}} \\
\bottomrule
\end{tabular}
\end{table}

%% file: table/ablations/table_encoder_decoder_layers.tex
\begin{table}[ht]
\centering
\small
\caption{
Effect of encoder--decoder depth on \longcounsel, \daic, and the complementary GPT-5.4-mini-generated dataset.
}
\label{tab:ablation-encoder-decoder-layers}
\begin{tabular}{ccccc}
\toprule
\multicolumn{2}{c}{\textbf{Layers}} & \multicolumn{3}{c}{\textbf{Overall MAE $\downarrow$}} \\
\cmidrule(lr){1-2}\cmidrule(l){3-5}
\textbf{Encoder} & \textbf{Decoder} & \longcounsel & \daic & \textsc{GPT-5.4-mini Comp.} \\
\midrule
1 & 4 & 2.6314 $\pm$ 0.0371 & 2.5125 $\pm$ 0.2239 & \textbf{3.8003 $\pm$ 0.0916} \\
2 & 2 & 2.6576 $\pm$ 0.0674 & 2.6274 $\pm$ 0.3043 & 3.8790 $\pm$ 0.2592 \\
2 & 4 & 2.6238 $\pm$ 0.0610~\textsuperscript{\ensuremath{\dagger}} & 2.6241 $\pm$ 0.2208 & 3.8447 $\pm$ 0.1181~\textsuperscript{\ensuremath{\dagger}} \\
2 & 6 & \textbf{2.6071 $\pm$ 0.0362} & 2.6053 $\pm$ 0.2492 & 3.8648 $\pm$ 0.0460 \\
3 & 4 & 2.6163 $\pm$ 0.0304 & \textbf{2.4708 $\pm$ 0.3892}~\textsuperscript{\ensuremath{\dagger}} & 3.8615 $\pm$ 0.1363 \\
\bottomrule
\end{tabular}
\end{table}

%% file: table/ablations/table_history_dropout.tex
\begin{table}[ht]
\centering
\small
\caption{
Effect of history dropout on \longcounsel and the complementary GPT-5.4-mini-generated dataset.
}
\label{tab:ablation-history-dropout}
\begin{tabular}{lcc}
\toprule
\multirow{2}{*}{\textbf{History dropout}} & \multicolumn{2}{c}{\textbf{Overall MAE $\downarrow$}} \\
\cmidrule(l){2-3}
& \longcounsel & \textsc{GPT-5.4-mini Comp.} \\
\midrule
0.0 & 2.6824 $\pm$ 0.0467 & \textbf{3.7898 $\pm$ 0.1085} \\
0.1 & \textbf{2.6238 $\pm$ 0.0610}~\textsuperscript{\ensuremath{\dagger}} & 3.8057 $\pm$ 0.0622 \\
0.2 & 2.6434 $\pm$ 0.0735 & 3.8447 $\pm$ 0.1181~\textsuperscript{\ensuremath{\dagger}} \\
0.3 & 2.6744 $\pm$ 0.0650 & 3.9462 $\pm$ 0.1312 \\
\bottomrule
\end{tabular}
\end{table}

%% file: table/ablations/table_max_context_sentences.tex
\begin{table}[ht]
\centering
\small
\caption{
Effect of the maximum number of retained client turns on \daic.
}
\label{tab:ablation-max-context-sentences}
\begin{tabular}{cc}
\toprule
\textbf{Maximum client turns} & \textbf{\daic Overall MAE $\downarrow$} \\
\midrule
20 & 2.7568 $\pm$ 0.2008 \\
40 & 2.7236 $\pm$ 0.3455 \\
80 & \textbf{2.4708 $\pm$ 0.3892}~\textsuperscript{\ensuremath{\dagger}} \\
384 & 2.5903 $\pm$ 0.2637 \\
\bottomrule
\end{tabular}
\end{table}

%% file: table/ablations/table_memory_mechanism.tex
\begin{table}[ht]
\centering
\small
\caption{
Effect of the memory mechanism on \longcounsel and the complementary GPT-5.4-mini-generated dataset.
}
\label{tab:ablation-memory-mechanism}
\begin{tabular}{lcc}
\toprule
\multirow{2}{*}{\textbf{Memory mechanism}} & \multicolumn{2}{c}{\textbf{Overall MAE $\downarrow$}} \\
\cmidrule(l){2-3}
& \longcounsel & \textsc{GPT-5.4-mini Comp.} \\
\midrule
No previous memory & 2.7097 $\pm$ 0.0514 & 3.8631 $\pm$ 0.0779 \\
Summary only & \textbf{2.6036 $\pm$ 0.0432} & 3.9369 $\pm$ 0.1281 \\
Summary + retrieval & 2.6238 $\pm$ 0.0610~\textsuperscript{\ensuremath{\dagger}} & \textbf{3.8447 $\pm$ 0.1181}~\textsuperscript{\ensuremath{\dagger}} \\
\bottomrule
\end{tabular}
\end{table}

%% file: table/ablations/table_memory_slots.tex
\begin{table}[ht]
\centering
\small
\caption{
Effect of memory capacity on \longcounsel and the complementary GPT-5.4-mini-generated dataset.
}
\label{tab:ablation-memory-slots}
\begin{tabular}{lcc}
\toprule
\multirow{2}{*}{\textbf{Memory slots}} & \multicolumn{2}{c}{\textbf{Overall MAE $\downarrow$}} \\
\cmidrule(l){2-3}
& \longcounsel & \textsc{GPT-5.4-mini Comp.} \\
\midrule
4 & \textbf{2.6238 $\pm$ 0.0610}~\textsuperscript{\ensuremath{\dagger}} & \textbf{3.8058 $\pm$ 0.0926} \\
8 & 2.6607 $\pm$ 0.0946 & 3.8801 $\pm$ 0.1487 \\
12 & 2.6617 $\pm$ 0.0416 & 3.8317 $\pm$ 0.0925 \\
16 & 2.6975 $\pm$ 0.0417 & 3.8447 $\pm$ 0.1181~\textsuperscript{\ensuremath{\dagger}} \\
\bottomrule
\end{tabular}
\end{table}

%% file: table/analysis/history_scope.tex
\begin{table}[ht]
\centering
\small
\caption{
Effect of history scope on \longcounsel and the complementary GPT-5.4-mini-generated dataset, with the previous-session compressor unchanged. Bold marks the lowest mean in each column.
}
\label{tab:history-scope}
\begin{tabular}{lcc}
\toprule
\multirow{2}{*}{\textbf{History scope}} & \multicolumn{2}{c}{\textbf{Overall MAE $\downarrow$}} \\
\cmidrule(l){2-3}
& \longcounsel & \textsc{GPT-5.4-mini Comp.} \\
\midrule
No previous memory & 2.7097 $\pm$ 0.0514 & 3.8631 $\pm$ 0.0779 \\
Previous 1 session & 2.6238 $\pm$ 0.0610 & 3.8447 $\pm$ 0.1181 \\
Previous 2 sessions & \textbf{2.5880 $\pm$ 0.0808} & \textbf{3.7559 $\pm$ 0.0977} \\
All history & 2.6111 $\pm$ 0.0742 & 3.7870 $\pm$ 0.1314 \\
\bottomrule
\end{tabular}
\end{table}

%% file: table/ablations/table_item_loss_total_item_mae.tex
\begin{table}[ht]
\centering
\small
\setlength{\tabcolsep}{2.5pt}
\caption{Effect of symptom-wise loss on total-score and item MAE. The weight $\lambda_{\mathrm{sym}}=0$ denotes aggregate-only supervision.}
\label{tab:ablation-item-loss-total-item-mae}
\begin{tabular}{@{}lcccc@{}}
\toprule
\multirow{2}{*}{\textbf{Training objective}} & \multicolumn{3}{c}{\textbf{Aggregate MAE $\downarrow$}} & \textbf{Symptom MAE $\downarrow$} \\
\cmidrule(lr){2-4}\cmidrule(l){5-5}
& \longcounsel & \daic & \makecell{\textsc{GPT-5.4-mini}\\Comp.} & \longcounsel \\
\midrule
$\lambda_{\mathrm{sym}}=0$ & 2.6481 $\pm$ 0.0370 & \textbf{2.4708 $\pm$ 0.3892}~\textsuperscript{\ensuremath{\dagger}} & 3.8447 $\pm$ 0.1181~\textsuperscript{\ensuremath{\dagger}} & 0.5674 $\pm$ 0.0041 \\
$\lambda_{\mathrm{sym}}=0.1$ & 2.6519 $\pm$ 0.0555 & 2.5600 $\pm$ 0.2717 & 3.8565 $\pm$ 0.1603 & 0.4847 $\pm$ 0.0206 \\
$\lambda_{\mathrm{sym}}=0.3$ & 2.6372 $\pm$ 0.0492 & 2.5785 $\pm$ 0.1866 & \textbf{3.7454 $\pm$ 0.1019} & 0.4310 $\pm$ 0.0125 \\
$\lambda_{\mathrm{sym}}=0.5$ & \textbf{2.6238 $\pm$ 0.0610}~\textsuperscript{\ensuremath{\dagger}} & 2.6937 $\pm$ 0.2892 & 3.7509 $\pm$ 0.0578 & 0.4251 $\pm$ 0.0124 \\
$\lambda_{\mathrm{sym}}=1.0$ & 2.6271 $\pm$ 0.0442 & 2.6477 $\pm$ 0.2494 & 3.8870 $\pm$ 0.1206 & \textbf{0.4178 $\pm$ 0.0131} \\
\bottomrule
\end{tabular}
\end{table}

%% file: table/analysis/fn_aware_comparison.tex
\begin{table}[ht]
\centering
\small
\caption{Screening comparison on the complementary dataset at a PHQ-8 threshold of $10$, with the lowest mean in each column in bold. Values are mean $\pm$ sample standard deviation over five seeds.}
\label{tab:fn-aware-comparison}
\begin{tabular}{lccc}
\toprule
\textbf{Method} & \textbf{FNR $\downarrow$} & \textbf{FPR $\downarrow$} & \textbf{MAE $\downarrow$} \\
\midrule
AIDA (now) & 0.7500 $\pm$ 0.0000 & \textbf{0.0778 $\pm$ 0.0037} & 4.0689 $\pm$ 0.0133 \\
AIDA (all) & 0.7300 $\pm$ 0.0187 & 0.0859 $\pm$ 0.0088 & 4.0948 $\pm$ 0.0243 \\
LMIQ (now) & 0.7040 $\pm$ 0.0167 & 0.0963 $\pm$ 0.0159 & 4.0095 $\pm$ 0.0566 \\
LMIQ (all) & 0.7220 $\pm$ 0.0349 & 0.0793 $\pm$ 0.0165 & 3.9908 $\pm$ 0.0513 \\
Milintsevich et al. & 0.6460 $\pm$ 0.0503 & 0.1481 $\pm$ 0.0183 & 4.1530 $\pm$ 0.0602 \\
EnsemBERT & 0.7560 $\pm$ 0.1704 & 0.0956 $\pm$ 0.0899 & 4.0038 $\pm$ 0.1396 \\
EnsemBERT (Qwen3-Emb) & 0.6620 $\pm$ 0.1182 & 0.1370 $\pm$ 0.0522 & 3.9184 $\pm$ 0.0753 \\
Lau et al. & 0.5880 $\pm$ 0.0638 & 0.1681 $\pm$ 0.0330 & 4.0689 $\pm$ 0.0865 \\
Lau et al. (Qwen3-Emb) & 0.6060 $\pm$ 0.0336 & 0.1756 $\pm$ 0.0279 & 4.0640 $\pm$ 0.1034 \\
\midrule
EmoTrack, $\lambda_{\mathrm{FN}}{=}0$ & 0.7080 $\pm$ 0.0694 & 0.1111 $\pm$ 0.0372 & \textbf{3.8447 $\pm$ 0.1181} \\
EmoTrack, $\lambda_{\mathrm{FN}}{=}2.5$ & \textbf{0.4220 $\pm$ 0.0853} & 0.2667 $\pm$ 0.0515 & 4.0370 $\pm$ 0.0691 \\
\bottomrule
\end{tabular}
\end{table}

%% file: table/analysis/symptom_bias.tex
\begin{table}[ht]
\centering
\small
\caption{
Per-symptom MAE and signed bias (prediction minus label) of EmoTrack on the 35-participant \daic development split.
}
\label{tab:symptom-bias}
\begin{tabular}{lcc}
\toprule
\textbf{PHQ-8 symptom} & \textbf{Item MAE $\downarrow$} & \textbf{Signed bias} \\
\midrule
Anhedonia & \textbf{0.4707} & $+0.2012$ \\
Depressed mood & 0.5191 & $-0.0871$ \\
Sleep & 0.7539 & $-0.3428$ \\
Fatigue & 0.6990 & $-0.4586$ \\
Appetite & 0.6931 & $-0.2293$ \\
Self-worth & 0.6618 & $-0.1438$ \\
Concentration & 0.6798 & $+0.0289$ \\
Psychomotor & 0.6226 & $+0.4554$ \\
\bottomrule
\end{tabular}
\end{table}

%% file: table/analysis/trajectory_strat.tex
\begin{table}[ht]
\centering
\small
\caption{Test MAE on \longcounsel by PHQ-8 trajectory type. Improving and worsening denote endpoint changes of $\leq-5$ and $\geq5$ points, fluctuating denotes a range $\geq5$ with absolute endpoint change $<5$, and stable denotes a range $<5$.}
\label{tab:trajectory-strat}
\begin{tabular}{lcccc}
\toprule
\multirow{2}{*}{\textbf{Method}} & \multicolumn{4}{c}{\textbf{Overall MAE $\downarrow$}} \\
\cmidrule(l){2-5}
& \makecell{Improving\\($n{=}93$)} & \makecell{Worsening\\($n{=}58$)} & \makecell{Stable\\($n{=}358$)} & \makecell{Fluctuating\\($n{=}229$)} \\
\midrule
AIDA (now) & 3.3936 $\pm$ 0.0128 & 3.7111 $\pm$ 0.0086 & 3.0385 $\pm$ 0.0100 & 3.3121 $\pm$ 0.0059 \\
AIDA (all) & 3.7581 $\pm$ 0.0095 & 3.8885 $\pm$ 0.0147 & 2.9432 $\pm$ 0.0035 & 3.3672 $\pm$ 0.0170 \\
LMIQ (now) & 3.1859 $\pm$ 0.0291 & 3.5164 $\pm$ 0.0301 & 2.6911 $\pm$ 0.0170 & 3.2094 $\pm$ 0.0219 \\
LMIQ (all) & 3.5495 $\pm$ 0.0895 & 3.7239 $\pm$ 0.0404 & 2.4882 $\pm$ 0.0186 & 3.2599 $\pm$ 0.0242 \\
Milintsevich et al. & 3.0254 $\pm$ 0.0660 & 3.2949 $\pm$ 0.0941 & 2.3030 $\pm$ 0.0437 & 3.0227 $\pm$ 0.0781 \\
Lau et al. & 2.9893 $\pm$ 0.0647 & 3.1561 $\pm$ 0.1006 & 2.2555 $\pm$ 0.0633 & 2.9053 $\pm$ 0.0943 \\
Lau et al. (Qwen3-Emb) & \textbf{2.9167 $\pm$ 0.0720} & \textbf{3.1270 $\pm$ 0.0570} & 2.2455 $\pm$ 0.0586 & \textbf{2.8740 $\pm$ 0.0620} \\
\midrule
EmoTrack & 2.9534 $\pm$ 0.1090 & 3.4494 $\pm$ 0.0635 & \textbf{2.1152 $\pm$ 0.0680} & 3.0760 $\pm$ 0.0917 \\
\bottomrule
\end{tabular}
\end{table}

%% file: table/methods_timing.tex
\begin{table}[ht]
\centering
\small
\caption{Inference time in single-GPU hours, using cached inputs for EmoTrack. Values are mean $\pm$ standard deviation.}
\label{tab:baseline-timing}
\begin{tabular}{lcc}
\toprule
\multirow{2}{*}{\textbf{Method}} 
& \multicolumn{2}{c}{\textbf{Inference}} \\
\cmidrule(l){2-3}
& \textbf{\longcounsel} & \textbf{\daic} \\
\midrule
AIDA (now) & $0.000\,\pm\, 0.000$ & $0.000\,\pm\, 0.000$ \\
LMIQ (now) & $0.011\,\pm\, 0.000$ & $0.000\,\pm\, 0.000$ \\
Milintsevich et al. & $5.74\,\pm\, 0.48$ & $0.330\,\pm\, 0.003$ \\
EnsemBERT & $0.004\,\pm\, 0.000$ & $0.001\,\pm\, 0.000$ \\
EnsemBERT (Qwen3-Emb) & $0.017\,\pm\, 0.001$ & $0.002\,\pm\, 0.000$ \\
Lau et al. & $7.23\,\pm\, 3.90$ & $0.254\,\pm\, 0.010$ \\
Lau et al. (Qwen3-Emb) & $23.72\,\pm\, 12.78$ & $0.636\,\pm\, 0.003$ \\
\midrule
{EmoTrack} (ours) & $0.029\,\pm\, 0.001$ & $0.004\,\pm\, 0.000$ \\
\bottomrule
\end{tabular}
\end{table}

%% file: section/limitation.tex
\section{Limitations}
\label{sec:limitations}

\paragraph{Real longitudinal clinical validation.}
Real multi-session counseling data with standardized session-level PHQ supervision are needed to validate longitudinal performance and the memory module in clinical settings. Our longitudinal evaluation uses synthetic counseling, while \daic provides real single-session interviews. We plan to extend the evaluation with clinical partners. The synthetic self-reports remain imperfect approximations of their generating states, and the dialogue audits measure selected aspects of similarity to RealCBT. The transition and trajectory analyses evaluate current-score error, leaving change-direction accuracy and complete-trajectory recovery untested.

\paragraph{Participant-level uncertainty.}
\label{subsec:participant-bootstrap}
The observed \daic improvement has substantial participant-level uncertainty. We average each participant's prediction over five seeds, compare absolute errors against AIDA, and bootstrap participants for a $95\%$ confidence interval. EmoTrack reduces MAE by $0.3324$ and improves absolute error for 28 of 47 participants ($59.6\%$), but the reduction interval is $[-0.1581,\ 0.8154]$ and a paired permutation test gives $p=0.1901$. This estimate scores seed-averaged predictions, while Table~\ref{tab:all-results} averages separately scored runs, yielding a mean difference of $0.3526$. The interval spanning zero leaves the population-level improvement uncertain.

\paragraph{History after severity changes.}
Previous-session memory can retain outdated evidence after changes of at least five PHQ points (Appendix~\ref{subsec:transitions}). The current gates do not fully suppress this effect. Future memory mechanisms should account explicitly for changes between sessions, with evaluation on real longitudinal data.

\paragraph{Upstream model and implementation dependence.}
Broader evaluation is needed to establish sensitivity to upstream model choice. The default configuration uses open-source embedding and clinical-feature models. Appendix~\ref{subsec:extractor-swap} also evaluates GPT-5.6 Luna as a proprietary extractor, under which EmoTrack retains a mean-MAE advantage over AIDA. This comparison covers one proprietary extractor and a separate reference configuration on one dataset. The cached-input timing in Table~\ref{tab:baseline-timing} excludes extraction and training, leaving end-to-end compute cost unmeasured. Baseline comparisons also depend on preprocessing, checkpoint selection, and evaluation protocols, as documented in Appendix~\ref{sec:add_experiment_setting}.

\paragraph{Item validity and clinical use.}
Item outputs have not been independently validated as symptom measurements, particularly under aggregate-only supervision. The clinical features are LLM-derived representations, and the PHQ-8 queries impose an architectural prior on total-score prediction. The public \daic test split lacks item labels, and the development-set audit finds opposing item biases that can cancel in the total score (Appendix~\ref{subsec:symptom-bias}). Symptom-level use requires item calibration and validation. The screening analysis in Section~\ref{subsec:optional-fn} and Appendix~\ref{subsec:fn-sweep} already evaluates the optional false-negative penalty, which increases false positives and score error as it reduces missed positives. Clinical use requires real-patient evaluation with application-specific operating costs, calibration, uncertainty estimation, privacy safeguards, and human review. Diagnostic and intervention uses remain unvalidated.

%% file: section/license.tex
\section{Asset licenses and data statement}
\label{sec:asset-license}

The EmoTrack source code is publicly available at \href{https://github.com/Xtra-Computing/EmoTrack}{github.com/Xtra-Computing/EmoTrack} under the Apache License 2.0.
The evaluation data comprise \longcounsel and two datasets from LongCounsel-8~\citep{li2026longcounsel}, LongCounsel-8-Luna and LongCounsel-8-GPT54-Mini. They are available at \href{https://huggingface.co/datasets/hiddensev/LongCounsel-8}{huggingface.co/datasets/hiddensev/LongCounsel-8} under the Creative Commons Attribution-NonCommercial 4.0 International (CC BY-NC 4.0) license. This license preserves the noncommercial condition stated by PSYCHE-D and PsychEval.
Table~\ref{tab:asset-licenses} gives a compact summary of new and existing asset terms. We cite all original assets and do not redistribute restricted third-party datasets.
\begin{table}[ht]
\centering
\small
\caption{Asset licenses, access terms, and intended use.}
\label{tab:asset-licenses}
\begin{tabular}{@{}p{0.22\linewidth}p{0.29\linewidth}p{0.41\linewidth}@{}}
\toprule
\textbf{Asset} & \textbf{Use} & \textbf{License/access terms} \\
\midrule
\textbf{EmoTrack code} & Training/evaluation implementation, baseline wrappers, and re-implementations. & Release repository: \href{\emotrackCodeURL}{\emotrackCodeURL}; Apache License 2.0. \\
\midrule
\textbf{\longcounsel} & Synthetic benchmark for transcript-based PHQ-8 prediction and longitudinal tracking. & CC BY-NC 4.0. Research use only; not for clinical diagnosis, autonomous triage, or claims about real individuals. \\
\midrule
\daic~\citep{Gratch2014TheDA} & Real single-session PHQ-8 benchmark. & USC ICT application-based access for academic/non-profit research; used under provider terms and not redistributed. \url{https://dcapswoz.ict.usc.edu/} \\
\midrule
NHANES DPQ\_L~\citep{cdc_nhanes_dpq_l_2024} & Symptom co-occurrence source for \longcounsel construction. & CDC/NCHS public-use data under the NCHS Data User Agreement; no re-identification or identifiable-data linkage. \url{https://www.cdc.gov/nchs/policy/data-user-agreement.html} \\
\midrule
PSYCHE-D~\citep{Makhmutova2021PredictingCI} & Aggregate depression-change trajectory source. & CC BY-NC 4.0. Used for trajectory reconstruction; raw source files are not redistributed. \url{https://zenodo.org/records/5085146} \\
\midrule
PsychEval~\citep{pan2026psycheval} & Case/background and session-stage structures. & CC BY-NC 4.0. The source is cited and is not redistributed as a separate raw asset. \url{https://github.com/ECNU-ICALK/PsychEval/blob/main/LICENSE} \\
\midrule
RealCBT~\citep{Wang2025FeelTD} & Dialogue-form calibration and similarity audit. & Provider/repository terms; cited and not redistributed as raw sessions. \url{https://gitlab.com/xiaoyi.wang/realcbt-dataset} \\
\midrule
Qwen models~\citep{qwen3.5} & Qwen3-Embedding-8B for embeddings and Qwen3.5-35B-A3B for feature extraction. & Apache License 2.0 as listed on the model cards. \url{https://huggingface.co/Qwen/Qwen3-Embedding-8B}; \url{https://huggingface.co/Qwen/Qwen3.5-35B-A3B} \\
\midrule
Baseline methods~\citep{lee2026interpretable,rosenman2024llm,lau2023automatic,milintsevich2023towards,ravenda2025transforming} & Reimplemented or adapted comparisons. & Original papers are cited; our released wrappers/reimplementations are under Apache License 2.0, and external author repositories are referenced but not redistributed. \\
\bottomrule
\end{tabular}
\end{table}

\paragraph{Access conditions and safeguards.}
The dataset card provides citation instructions, edition descriptions, split manifests, and access instructions. Users request access on Hugging Face and authenticate after approval to download these datasets. The code release includes license files and reproduction instructions.
The dataset is intended for research. Users must respect upstream access restrictions and avoid claims about real individuals based on synthetic records.

%% file: checklist.tex
\section*{NeurIPS Paper Checklist}

\begin{enumerate}

\item {\bf Claims}
    \item[] Question: Do the main claims made in the abstract and introduction accurately reflect the paper's contributions and scope?
    \item[] Answer: \answerYes{}.
    \item[] Justification: The abstract and Section~\ref{sec:intro} state the method, evaluation settings, and main MAE claim, which are supported by Sections~\ref{sec:approach} and \ref{sec:experiment}. Appendix~\ref{apdx:longcounsel} documents the longitudinal dataset. The scope is bounded by the limitations discussed in Appendix~\ref{sec:limitations}.

\item {\bf Limitations}
    \item[] Question: Does the paper discuss the limitations of the work performed by the authors?
    \item[] Answer: \answerYes{}.
    \item[] Justification: Appendix~\ref{sec:limitations} discusses the need for real longitudinal clinical validation, the use of open-source upstream models, and the distinction between PHQ-8 prediction and deployable clinical decision support.

\item {\bf Theory assumptions and proofs}
    \item[] Question: For each theoretical result, does the paper provide the full set of assumptions and a complete (and correct) proof?
    \item[] Answer: \answerNA{}.
    \item[] Justification: The paper does not present theoretical results or proofs. The equations in Section~\ref{sec:approach} define the model and training objective.

	\item {\bf Experimental result reproducibility}
    \item[] Question: Does the paper fully disclose all the information needed to reproduce the main experimental results of the paper to the extent that it affects the main claims and/or conclusions of the paper (regardless of whether the code and data are provided or not)?
    \item[] Answer: \answerYes{}.
	    \item[] Justification: Sections~\ref{sec:approach} and \ref{sec:experiment} and Appendices~\ref{apdx:emotrack-algorithm-detail}, \ref{sec:add_experiment_setting}, \ref{apdx:add_experiment}, and \ref{sec:asset-license} provide the model, splits, metrics, baselines, hyperparameters, asset terms, and release plan. Section~\ref{subsec:experiment-settings} links the evaluation data, and Appendix~\ref{sec:asset-license} describes data access and the planned code release.

\item {\bf Open access to data and code}
    \item[] Question: Does the paper provide open access to the data and code, with sufficient instructions to faithfully reproduce the main experimental results, as described in supplemental material?
    \item[] Answer: \answerYes{}.
	    \item[] Justification: Appendix~\ref{sec:asset-license} documents the code and data release under Apache 2.0 and CC BY-NC 4.0, respectively, and provides the GitHub code URL and dataset URL together with access conditions.

\item {\bf Experimental setting/details}
    \item[] Question: Does the paper specify all the training and test details (e.g., data splits, hyperparameters, how they were chosen, type of optimizer) necessary to understand the results?
    \item[] Answer: \answerYes{}.
    \item[] Justification: Section~\ref{subsec:experiment-settings} specifies datasets, splits, upstream models, feature extraction, evaluation protocol, and baselines. Appendix~\ref{sec:add_experiment_setting} provides hyperparameters and training details.

\item {\bf Experiment statistical significance}
    \item[] Question: Does the paper report error bars suitably and correctly defined or other appropriate information about the statistical significance of the experiments?
    \item[] Answer: \answerYes{}.
    \item[] Justification: Section~\ref{subsec:experiment-settings} states that results are reported as five-seed means with standard deviations, and the result tables report mean $\pm$ sample standard deviation. The seed variability is described through matched-seed feature extraction and downstream training. Appendix~\ref{subsec:participant-bootstrap} adds a participant-level bootstrap interval and paired permutation test for the main \daic comparison.

\item {\bf Experiments compute resources}
    \item[] Question: For each experiment, does the paper provide sufficient information on the computer resources (type of compute workers, memory, time of execution) needed to reproduce the experiments?
    \item[] Answer: \answerYes{}.
	    \item[] Justification: Appendix~\ref{sec:add_experiment_setting} reports the local server system, GPU, CPU, and memory. Appendix~\ref{subsec:runtime-compute-cost} and Table~\ref{tab:baseline-timing} provide detailed per-method inference time and single-GPU-hour estimates for the main comparison runs.
    
\item {\bf Code of ethics}
    \item[] Question: Does the research conducted in the paper conform, in every respect, with the NeurIPS Code of Ethics \url{https://neurips.cc/public/EthicsGuidelines}?
    \item[] Answer: \answerYes{}.
    \item[] Justification: Appendix~\ref{sec:limitations} discusses clinical validation, privacy safeguards, and human-in-the-loop review. Appendix~\ref{sec:asset-license} documents asset terms, intended use, and release conditions.

\item {\bf Broader impacts}
    \item[] Question: Does the paper discuss both potential positive societal impacts and negative societal impacts of the work performed?
    \item[] Answer: \answerYes{}.
    \item[] Justification: Section~\ref{sec:intro} motivates potential use for monitoring depression severity and supporting timely human review. Appendix~\ref{sec:limitations} discusses risks around clinical validation, privacy safeguards, false negatives, and human-in-the-loop deployment.
    
\item {\bf Safeguards}
    \item[] Question: Does the paper describe safeguards that have been put in place for responsible release of data or models that have a high risk for misuse (e.g., pre-trained language models, image generators, or scraped datasets)?
    \item[] Answer: \answerYes{}.
    \item[] Justification: Appendix~\ref{sec:asset-license} specifies release licenses, intended use, access conditions, no redistribution of restricted upstream assets, and safeguards against clinical misuse of the synthetic mental-health benchmark. Appendix~\ref{sec:limitations} further discusses deployment safeguards.

\item {\bf Licenses for existing assets}
    \item[] Question: Are the creators or original owners of assets (e.g., code, data, models), used in the paper, properly credited and are the license and terms of use explicitly mentioned and properly respected?
    \item[] Answer: \answerYes{}.
    \item[] Justification: Existing datasets, models, and baselines are cited throughout Sections~\ref{sec:related} and \ref{sec:experiment} and Appendix~\ref{apdx:longcounsel}. Appendix~\ref{sec:asset-license} summarizes the applicable licenses or provider access terms for \daic, NHANES/DPQ, PSYCHE-D, RealCBT, PsychEval-derived assets, Qwen models, and baseline implementations.

\item {\bf New assets}
    \item[] Question: Are new assets introduced in the paper well documented and is the documentation provided alongside the assets?
    \item[] Answer: \answerYes{}.
    \item[] Justification: \longcounsel is documented in Section~\ref{subsec:experiment-settings} and Appendix~\ref{apdx:longcounsel}. Appendix~\ref{sec:asset-license} adds a data statement covering license, intended use, limitations, access conditions, and release safeguards.

\item {\bf Crowdsourcing and research with human subjects}
    \item[] Question: For crowdsourcing experiments and research with human subjects, does the paper include the full text of instructions given to participants and screenshots, if applicable, as well as details about compensation (if any)? 
    \item[] Answer: \answerNA{}.
    \item[] Justification: The paper does not report new crowdsourcing experiments or new human-subject data collection. It uses synthetic data and existing datasets, whose asset terms should be documented under the licenses and new-assets checklist items.

\item {\bf Institutional review board (IRB) approvals or equivalent for research with human subjects}
    \item[] Question: Does the paper describe potential risks incurred by study participants, whether such risks were disclosed to the subjects, and whether Institutional Review Board (IRB) approvals (or an equivalent approval/review based on the requirements of your country or institution) were obtained?
    \item[] Answer: \answerNA{}.
    \item[] Justification: The paper does not conduct new human-subject or crowdsourcing studies. Existing human-derived corpora are used as external assets, and their access conditions and original approvals should be covered through the asset documentation.

\item {\bf Declaration of LLM usage}
    \item[] Question: Does the paper describe the usage of LLMs if it is an important, original, or non-standard component of the core methods in this research? Note that if the LLM is used only for writing, editing, or formatting purposes and does \emph{not} impact the core methodology, scientific rigor, or originality of the research, declaration is not required.
    \item[] Answer: \answerYes{}.
    \item[] Justification: Section~\ref{sec:approach} describes LLM-extracted clinical features as part of EmoTrack, and Section~\ref{subsec:experiment-settings} identifies the Qwen models used for feature extraction and embeddings. Appendix~\ref{apdx:longcounsel} describes LLM-based synthetic dialogue construction.

\end{enumerate}